\documentclass[11pt]{article}
\usepackage{ACL2023}

\makeatletter
\AtBeginDocument{%
  \@twocolumnfalse            % tell LaTeX we're not in two-column mode
  \let\orig@twocolumn\twocolumn
  \renewcommand{\twocolumn}[1][]{\onecolumn} % any attempt to go two-col becomes one-col
  \onecolumn                  % start in one column
}
\makeatother

\usepackage{microtype}
\usepackage{xcolor}
\usepackage{graphicx}
\usepackage{booktabs}
\usepackage{array}
\usepackage{tabularx}
\usepackage{longtable}
\usepackage{multirow}
\usepackage{float}
\usepackage{wrapfig}
\usepackage{enumitem}
\usepackage{amsmath,amssymb,breqn}
\usepackage{algorithm}
\usepackage{algpseudocode}
\usepackage{framed}
\usepackage{comment}
\usepackage{natbib}
\usepackage{multibib}
\usepackage{pifont}
\usepackage{arydshln}
\usepackage{tikz}
\usetikzlibrary{trees,shapes,shapes.geometric,arrows,decorations.markings,positioning,arrows.meta}
\usepackage[many]{tcolorbox}
\usepackage{subcaption}
\usepackage{varwidth}
\usepackage{setspace}
\usepackage{extsizes}
\usepackage{cuted}
\usepackage{flushend}
\usepackage{dblfloatfix}
\usepackage{fixltx2e}
\usepackage{soul}
\usepackage{fontspec}
\usepackage{hyperref}

\definecolor{darkblue}{rgb}{0,0,0.5}
\hypersetup{colorlinks=true, citecolor=darkblue, linkcolor=darkblue, urlcolor=darkblue}
\usepackage[capitalise,nameinlink]{cleveref}
\usepackage{wasysym} % for \bigstar
\newcommand{\benchmarkname}{\textsc{SleepWalk}}
\newcommand{\scenegenmodel}{Hunyuan3D-3.0}
\newcommand{\instrgenmodel}{Qwen3-8B-VL}

\newcommand{\numenvs}{2,472}
\newcommand{\instrperscene}{9}

\usepackage[many]{tcolorbox}

\newtcolorbox{defin}{colback=Teal!5!White,enhanced,title=Alignment Faking: Bayesian–Stackelberg Equilibria,	attach boxed title to top left={xshift=-4mm},boxrule=0pt,after skip=1cm,before skip=1cm,right skip=0cm,breakable,fonttitle=\bfseries,toprule=0pt,bottomrule=0pt,rightrule=0pt,leftrule=3pt,arc=0mm,skin=enhancedlast jigsaw,sharp corners,colframe=Teal!55!black,colbacktitle=Teal!55!black,boxed title style={
		frame code={ 
			\fill[Teal!25!black](frame.south west)--(frame.north west)--(frame.north east)--([xshift=3mm]frame.east)--(frame.south east)--cycle;
			\draw[line width=1mm,Teal!25!black]([xshift=2mm]frame.north east)--([xshift=5mm]frame.east)--([xshift=2mm]frame.south east);
			\draw[line width=1mm,Teal!25!black]([xshift=5mm]frame.north east)--([xshift=8mm]frame.east)--([xshift=5mm]frame.south east);
			\fill[Teal!25!black](frame.south west)--+(4mm,-2mm)--+(4mm,2mm)--cycle;
		}
	}
}
\usetikzlibrary{shapes.geometric, arrows}
\usetikzlibrary{decorations.markings}

\definecolor{first}{RGB}{210,255,140}
\definecolor{second}{RGB}{136, 162, 190}
\definecolor{third}{RGB}{129, 222, 228}
\definecolor{fourth}{RGB}{132, 84, 246}
\definecolor{fifth}{RGB}{250, 223, 112}
\definecolor{sixth}{RGB}{203, 193, 172}
\definecolor{seventh}{RGB}{88, 112, 246}
\definecolor{eighth}{RGB}{245, 192, 106}
\definecolor{nine}{RGB}{171, 162, 111}
\definecolor{ten}{RGB}{217, 217, 217}

\definecolor{paired-light-blue}{RGB}{198, 219, 239}
\definecolor{paired-dark-blue}{RGB}{49, 130, 188}
\definecolor{paired-light-orange}{RGB}{251, 208, 162}
\definecolor{paired-dark-orange}{RGB}{230, 85, 12}
\definecolor{paired-light-green}{RGB}{199, 233, 193}
\definecolor{paired-dark-green}{RGB}{49, 163, 83}
\definecolor{paired-light-purple}{RGB}{218, 218, 235}
\definecolor{paired-dark-purple}{RGB}{117, 107, 176}
\definecolor{paired-light-gray}{RGB}{217, 217, 217}
\definecolor{paired-dark-gray}{RGB}{99, 99, 99}
\definecolor{paired-light-pink}{RGB}{222, 158, 214}
\definecolor{paired-dark-pink}{RGB}{123, 65, 115}
\definecolor{paired-light-red}{RGB}{231, 150, 156}
\definecolor{paired-dark-red}{RGB}{131, 60, 56}
\definecolor{paired-light-yellow}{RGB}{231, 204, 149}
\definecolor{paired-dark-yellow}{RGB}{141, 109, 49}
\definecolor{Teal}{RGB}{0, 50, 50}
\definecolor{White}{RGB}{250, 250, 250}
\definecolor{bg1}{HTML}{FF9966}
\definecolor{bg2}{HTML}{CCE5FF}
\definecolor{bg3}{HTML}{FFCC99}
\definecolor{bg4}{HTML}{FFC107}
\definecolor{bg5}{HTML}{FFCCCC}
\definecolor{bg6}{HTML}{D5E8D4}
\definecolor{bg7}{HTML}{eeeeee}
\definecolor{bg8}{HTML}{cdeb8b}
\definecolor{bg9}{HTML}{dae8fc}
\definecolor{bg10}{HTML}{a2e6eb}
\definecolor{bg31}{HTML}{FFCDD2} % light pink
\definecolor{bg32}{HTML}{F8BBD0}
\definecolor{bg33}{HTML}{E1BEE7} % lavender
\definecolor{bg34}{HTML}{D7CCC8} % light tan
\definecolor{bg35}{HTML}{B2DFDB} % light teal
\definecolor{bg36}{HTML}{A5D6A7} % light green
\definecolor{bg37}{HTML}{FFF9C4} %light yellow
\definecolor{bg38}{HTML}{FFECB3} % peach
\definecolor{bg111}{HTML}{CB6843}
\definecolor{bg112}{HTML}{D77C5C}
\definecolor{bg113}{HTML}{E28E6E}
\definecolor{bg114}{HTML}{E89F7D}
\definecolor{bg115}{HTML}{EDAE8A}
\definecolor{bg116}{HTML}{F0BA95}
\definecolor{bg117}{HTML}{F3C29F}
\definecolor{bg118}{HTML}{F6CCAA}
\definecolor{bg119}{HTML}{F8D5B3}
\definecolor{bg120}{HTML}{FADCBD}
\definecolor{bg121}{HTML}{FCE6C7}
\definecolor{bg39}{HTML}{FFE0B2} % apricot
\definecolor{bg40}{HTML}{3CB371} % blush pink

\definecolor{bg43}{HTML}{ffe5d9}
\definecolor{bg15}{HTML}{7FFFD4}
\definecolor{bg17}{HTML}{F0FFFF}
\definecolor{bg18}{HTML}{F5FFFA}
\definecolor{bg19}{HTML}{F8F8FF}
\definecolor{bg20}{HTML}{FFFFFF}
\definecolor{bg21}{HTML}{E1F5FE}
\definecolor{bg22}{HTML}{B3E5FC}
\definecolor{bg23}{HTML}{81D4FA}
\definecolor{bg24}{HTML}{4FC3F7}
\definecolor{bg25}{HTML}{29B6F6}
\definecolor{bg26}{HTML}{03A9F4}
\definecolor{bg27}{HTML}{039BE5}
\definecolor{bg28}{HTML}{0288D1}
\definecolor{bg29}{HTML}{0277BD}
\definecolor{bg30}{HTML}{01579B}
\definecolor{bg16}{HTML}{FFCC99} 
\definecolor{pg51}{HTML}{E8F5E9} % pale green
\definecolor{pg52}{HTML}{C8E6C9} % honeydew green
\definecolor{pg53}{HTML}{B9F6CA} % light mint green
\definecolor{pg54}{HTML}{A9DFBF} % pale sage green
\definecolor{pg55}{HTML}{BCF5A6} % lemon green
\definecolor{pg56}{HTML}{BEF1CE} % seashell green
\definecolor{pg57}{HTML}{CEF6EC} % icy green
\definecolor{pg58}{HTML}{B7F0B1} % feijoa green
\definecolor{pg59}{HTML}{B1F2B5} % pastel light green
\definecolor{pg60}{HTML}{9DF3C4} % greenish cyan
\definecolor{pg61}{HTML}{DEF7E0} % pale green
\definecolor{pg62}{HTML}{E8F8DC} % greenish beige
\definecolor{pg63}{HTML}{EBF7E7} % seafoam green
\definecolor{pg64}{HTML}{F0FDF4} % pale turquoise
\definecolor{pg65}{HTML}{F1FEE7} % mint cream
\definecolor{pg66}{HTML}{F7FFF6} % foam green
\definecolor{pg67}{HTML}{FCFFE7} % pale spring bud
\definecolor{pg68}{HTML}{F4FFD2} % light lime green
\definecolor{pg69}{HTML}{EEFFE2} % tea green
\definecolor{pg70}{HTML}{E3FDF5} % tropical green
\definecolor{connect-color}{RGB}{0,0,0}
\definecolor{middle-color}{RGB}{255,255,255}
\definecolor{leaf-color}{RGB}{173,216,230}
\definecolor{line-color}{RGB}{25,25,112}
\definecolor{soothingPurple}{RGB}{195, 160, 201}
\definecolor{hidden-draw}{RGB}{20,68,106}
\definecolor{hidden-pink}{RGB}{255,245,247}
\definecolor{dark-red}{RGB}{233, 150, 122}
\definecolor{light-red}{RGB}{255,182,193}
\definecolor{medium-red}{RGB}{205,92,92}
\definecolor{light-yellow}{RGB}{255, 239, 153}
\definecolor{light-blue}{RGB}{173, 216, 230}
\definecolor{paired-light-yellow}{HTML}{FFFF88}
\definecolor{paired-light-blue}{HTML}{CCE5FF}
\definecolor{paired-light-orange}{HTML}{FFCC99}
\definecolor{paired-dark-yellow}{HTML}{FFF2CC}
\definecolor{paired-light-pink}{HTML}{FFCCCC}
\definecolor{paired-cyan}{HTML}{D5E8D4}
\definecolor{paired-gray}{HTML}{eeeeee}
\definecolor{paired-green}{HTML}{cdeb8b}
\definecolor{paired-blue}{HTML}{dae8fc}
\definecolor{paired-dark-cyan}{HTML}{a2e6eb}
\definecolor{paired-dark-pink}{HTML}{e7b2d2}
\definecolor{paired-purple}{HTML}{9999ff}
\definecolor{paired-pink}{HTML}{cc99ff}
\definecolor{paired-orange}{HTML}{ffcc99}

\definecolor{a1}{RGB}{241,233,191}
\definecolor{a2}{RGB}{255,241,218}

\definecolor{a3}{RGB}{255,239,213}
\definecolor{a4}{RGB}{250,235,215}
\definecolor{a5}{RGB}{255,239,219}
\definecolor{a6}{RGB}{255,246,225}
\definecolor{a7}{RGB}{246,227,201}
\definecolor{a8}{RGB}{254,235,226}
\definecolor{a9}{RGB}{247,220,111}
\definecolor{a10}{RGB}{199,211,189}
\definecolor{a11}{RGB}{209,196,233}
\definecolor{a12}{RGB}{214,234,248}
\definecolor{a13}{RGB}{232,245,233}
\definecolor{a14}{RGB}{237,248,177}
\definecolor{a15}{RGB}{255,228,225}
\definecolor{a16}{RGB}{255,228,181}
\definecolor{a17}{RGB}{255,222,173}
\definecolor{a18}{RGB}{255,218,185}
\definecolor{a19}{RGB}{255,203,164}
\definecolor{a20}{RGB}{247,202,201}

\definecolor{a21}{RGB}{241,254,255}
\definecolor{a22}{RGB}{230,252,252}
\definecolor{a23}{RGB}{179,236,255}
\definecolor{a24}{RGB}{174,226,249}
\definecolor{a25}{RGB}{208,234,246}
\definecolor{a26}{RGB}{189,226,219}
\definecolor{a27}{RGB}{177,204,201}

\definecolor{a28}{RGB}{216,195,216}
\definecolor{a29}{RGB}{195,155,211}
\definecolor{a30}{RGB}{208,152,223}
\definecolor{a31}{RGB}{255,183,209}
\definecolor{a32}{RGB}{255,167,209}
\definecolor{a33}{RGB}{254,235,167}
\definecolor{a34}{RGB}{255,222,137}
\definecolor{a35}{RGB}{254,180,154}
\definecolor{a36}{RGB}{247,148,161}
\definecolor{a37}{RGB}{239,154,154}
\definecolor{a38}{RGB}{255,130,171}
\definecolor{a39}{RGB}{255,105,180}
\definecolor{a40}{RGB}{251,142,172}

\newtcolorbox{societal_harm}{
  colback=soothingPurple, % Background color
  colframe=black, % Border color
  boxrule=0pt,% Border width
  enhanced,
  title=Societal harm,
  attach boxed title to top right={yshift=-3mm},
  fonttitle=\bfseries,
  toprule=1pt,
  bottomrule=1pt,
  rightrule=1pt,
  leftrule=1pt,
  arc=1mm
}

\newtcolorbox{privacy_violation}{
  colback=soothingPurple, % Background color
  colframe=black, % Border color
  boxrule=0pt,% Border width
  enhanced,
  title=Privacy Violation,
  attach boxed title to top right={yshift=-3mm},
  fonttitle=\bfseries,
  toprule=1pt,
  bottomrule=1pt,
  rightrule=1pt,
  leftrule=1pt,
  arc=1mm
}

\newtcolorbox{disinformation_deception}{
  colback=soothingPurple, % Background color
  colframe=black, % Border color
  boxrule=0pt,% Border width
  enhanced,
  title=Disinformation \& Deception,
  attach boxed title to top right={yshift=-3mm},
  fonttitle=\bfseries,
  toprule=1pt,
  bottomrule=1pt,
  rightrule=1pt,
  leftrule=1pt,
  arc=1mm
}

\newtcolorbox{answer_disparity}{
  colback=soothingPurple, % Background color
  colframe=black, % Border color
  boxrule=0pt,% Border width
  enhanced,
  title=Answer disparity,
  attach boxed title to top right={yshift=-3mm},
  fonttitle=\bfseries,
  toprule=1pt,
  bottomrule=1pt,
  rightrule=1pt,
  leftrule=1pt,
  arc=1mm
}

\newtcolorbox{wrong_classification}{
  colback=soothingPurple, % Background color
  colframe=black, % Border color
  boxrule=0pt,% Border width
  enhanced,
  title=Wrong classification,
  attach boxed title to top right={yshift=-3mm},
  fonttitle=\bfseries,
  toprule=1pt,
  bottomrule=1pt,
  rightrule=1pt,
  leftrule=1pt,
  arc=1mm
}

\newtcolorbox{goal_hijacking}{
  colback=soothingPurple, % Background color
  colframe=black, % Border color
  boxrule=0pt,% Border width
  enhanced,
  title=Goal hijacking,
  attach boxed title to top right={yshift=-3mm},
  fonttitle=\bfseries,
  toprule=1pt,
  bottomrule=1pt,
  rightrule=1pt,
  leftrule=1pt,
  arc=1mm
}

\newtcolorbox{control_generation}{
  colback=soothingPurple, % Background color
  colframe=black, % Border color
  boxrule=0pt,% Border width
  enhanced,
  title=Control generation,
  attach boxed title to top right={yshift=-3mm},
  fonttitle=\bfseries,
  toprule=1pt,
  bottomrule=1pt,
  rightrule=1pt,
  leftrule=1pt,
  arc=1mm
}

\newtcolorbox{prompt_leaking}{
  colback=soothingPurple, % Background color
  colframe=black, % Border color
  boxrule=0pt,% Border width
  enhanced,
  title=Prompt leaking,
  attach boxed title to top right={yshift=-3mm},
  fonttitle=\bfseries,
  toprule=1pt,
  bottomrule=1pt,
  rightrule=1pt,
  leftrule=1pt,
  arc=1mm
}

\usepackage{lipsum}
\usepackage{tikz}
\usetikzlibrary{trees,shapes}

\usepackage{times}
\usepackage{latexsym}

\usepackage[T5]{fontenc}
\usepackage[utf8]{inputenc}

\usepackage{microtype}

\usepackage{inconsolata}
\usepackage{soul}

\usepackage{inconsolata}
\usepackage{algorithm}
\usepackage{algpseudocode}
\usepackage{amsmath,amssymb}
\usepackage{soul}
\usepackage{tikz}

\tikzset{rndblock/.style={rounded corners,rectangle,draw,scale=0.8,outer sep=0pt}}

\usetikzlibrary{shapes.geometric}
\usepackage{framed}
\usepackage{enumitem}
\newlist{RQ}{enumerate}{1}
\setlist[RQ]{label=\textbf{RQ\,\arabic*},ref={RQ\,\arabic*}}
\usepackage{comment}
\usepackage{natbib}
\usepackage{multibib}
\makeatletter
\usepackage{booktabs}
\usepackage[inkscapeformat=png]{svg}
\usepackage{graphicx}
\usepackage{caption}
\usepackage{subcaption}
\usepackage{tabularx}
\usepackage{soul}
\usepackage{float}
\usepackage{enumitem}
\usepackage{pifont}
\usepackage{arydshln}
\usepackage{lipsum}

\usepackage[many]{tcolorbox}

\usetikzlibrary{shapes.geometric, arrows}
\usetikzlibrary{decorations.markings}

\usepackage{fancybox}

\usepackage{hyperref}
 \definecolor{darkblue}{rgb}{0, 0, 0.5}
  \hypersetup{colorlinks=true, citecolor=darkblue, linkcolor=darkblue, urlcolor=darkblue}

\definecolor{vgreen}{HTML}{60A917}
\definecolor{vred}{HTML}{CE3A29}

\usepackage{xstring}
\usepackage{longtable}

\usepackage{tabularray}

\DefTblrTemplate{firsthead,middlehead,lasthead}{default}{
}
\DefTblrTemplate{firstfoot}{default}{
  \UseTblrTemplate{contfoot}{default}
  \UseTblrTemplate{caption}{default}
}
\DefTblrTemplate{middlefoot}{default}{
  \UseTblrTemplate{contfoot}{default}
  \UseTblrTemplate{capcont}{default}
}
\DefTblrTemplate{lastfoot}{default}{
  \UseTblrTemplate{note}{default}
  \UseTblrTemplate{remark}{default}
  \UseTblrTemplate{capcont}{default}
}

\newcolumntype{P}[1]{>{\centering\arraybackslash}p{#1}}
\usepackage{color}
\tcbuselibrary{skins}

\usepackage[export]{adjustbox} % for the valign option

\usepackage{setspace}
\usepackage[capitalise,nameinlink]{cleveref}

\crefname{section}{Sec.}{Sec.}

\usepackage{microtype}
\usepackage{hyperref}
\usepackage{graphicx}
\usepackage{comment}
\usepackage{amsmath}
\usepackage{amssymb}
\usepackage{algorithm}
\usepackage{algpseudocode}
\usepackage{colortbl}
\usepackage[export]{adjustbox} % for the valign option
\usepackage{varwidth}
\usepackage{enumitem}
\setlist{leftmargin=1mm}
\usepackage{pifont}
\usepackage{booktabs}
\usepackage{multirow}
\usepackage{subcaption}
\usepackage{resizegather}
\usepackage{breqn}
\usepackage[capitalise]{cleveref}
\usepackage{graphicx}
\usepackage{tikz}
\usetikzlibrary{shapes.geometric, arrows}
\usetikzlibrary{decorations.markings}
\usepackage{soul}
\usepackage{wrapfig,graphicx,lipsum}% http://ctan.org/pkg/{wrapfig,graphicx,lipsum}
\usepackage{extsizes}
\usepackage{cuted}
\usepackage{flushend}
\usepackage{float}
\usepackage{changepage,threeparttable}
\usepackage{setspace}
\usepackage{caption}
\usepackage{booktabs}
\usepackage{dblfloatfix} 
\usepackage{fixltx2e}
\usepackage[normalem]{ulem}
\usepackage{microtype}
\usepackage{hyperref}
\usepackage{url}
\usepackage{booktabs}
\usepackage{graphicx}
\usepackage{amsmath}
\usepackage{graphicx}
\usepackage{subcaption}
\usepackage{todonotes}
\usepackage{amsmath,amssymb}
\usepackage{enumitem}
\usepackage{listings}

\usepackage{fvextra}
\DefineVerbatimEnvironment{Verbatim}{Verbatim}{
  breaklines=true,
  breakanywhere=true
}

\usepackage{environ}
\usepackage{soul}
\usepackage{float}
\usepackage{enumitem}
\usepackage{pifont}
\usepackage{arydshln}
\usepackage{lipsum}

\usepackage[many]{tcolorbox}

\usetikzlibrary{shapes.geometric, arrows}
\usetikzlibrary{decorations.markings}

\usepackage{fancybox}
\usepackage{hyperref}
 \definecolor{darkblue}{rgb}{0, 0, 0.5}
  \hypersetup{colorlinks=true, citecolor=darkblue, linkcolor=darkblue, urlcolor=darkblue}

\definecolor{vgreen}{HTML}{60A917}
\definecolor{vred}{HTML}{CE3A29}

\usepackage{xstring}
\usepackage{longtable}

\usepackage{tabularray}

\DefTblrTemplate{firsthead,middlehead,lasthead}{default}{
}
\DefTblrTemplate{firstfoot}{default}{
  \UseTblrTemplate{contfoot}{default}
  \UseTblrTemplate{caption}{default}
}
\DefTblrTemplate{middlefoot}{default}{
  \UseTblrTemplate{contfoot}{default}
  \UseTblrTemplate{capcont}{default}
}
\DefTblrTemplate{lastfoot}{default}{
  \UseTblrTemplate{note}{default}
  \UseTblrTemplate{remark}{default}
  \UseTblrTemplate{capcont}{default}
}

\usepackage{color}
\tcbuselibrary{skins}

\usepackage[export]{adjustbox} % for the valign option

\usepackage{setspace}
\usepackage[capitalise,nameinlink]{cleveref}

\crefname{section}{Sec.}{Sec.}

\usepackage{microtype}
\usepackage{hyperref}
\usepackage{graphicx}
\usepackage{comment}
\usepackage{amsmath}
\usepackage{amssymb}
\usepackage{algorithm}
\usepackage{algpseudocode}
\usepackage{colortbl}
\usepackage[export]{adjustbox} % for the valign option
\usepackage{varwidth}
\usepackage{enumitem}
\setlist{leftmargin=1mm}
\usepackage{pifont}
\usepackage{booktabs}
\usepackage{multirow}
\usepackage{subcaption}
\usepackage{resizegather}
\usepackage{breqn}
\usepackage[capitalise]{cleveref}
\usepackage{graphicx}
\usepackage{tikz}
\usetikzlibrary{shapes.geometric, arrows}
\usetikzlibrary{decorations.markings}
\usepackage{soul}
\usepackage{wrapfig,graphicx,lipsum}% http://ctan.org/pkg/{wrapfig,graphicx,lipsum}
\usepackage{extsizes}
\usepackage{cuted}
\usepackage{flushend}
\usepackage{float}
\usepackage{changepage,threeparttable}
\usepackage{setspace}
\usepackage{caption}
\usepackage{booktabs}
\usepackage{dblfloatfix} 
\usepackage{fixltx2e}
\usepackage[normalem]{ulem}

\usepackage{tcolorbox}
\usepackage{amsmath}
\usepackage{amssymb}
\usepackage{caption}

\usepackage{environ}

\newlength{\myl}
\expandafter\let\expandafter\origequation\csname equation*\endcsname
\expandafter\let\expandafter\endorigequation\csname endequation*\endcsname
\long\def\[#1\]{\begin{equation*}#1\end{equation*}}
\RenewEnviron{equation*}{
  \settowidth{\myl}{$\displaystyle\BODY$} % calculate width and save as \myl
  \origequation
    \ifdim\myl>\linewidth
      \resizebox{\linewidth}{!}{$\displaystyle\BODY$}% \myl > \linewidth
    \else
      \BODY % \myl <= \linewidth
    \fi
  \endorigequation
}

\makeatletter
\newcommand{\DrawLine}{%
  \begin{tikzpicture}
  \path[use as bounding box] (0,0) -- (\linewidth,0);
  \draw[color=blue!75!black,dashed,dash phase=.5pt]
        (0-\kvtcb@leftlower-\kvtcb@boxsep,0)--
        (\linewidth+\kvtcb@rightlower+\kvtcb@boxsep,0);
  \end{tikzpicture}%
  }
\makeatother

\usepackage{pgfplots}
\usepackage{tikz}
\usetikzlibrary{positioning, arrows.meta}
\usepackage{longtable}
\usepackage{booktabs}
\usepackage{array}
\usepackage{adjustbox}
\usepackage{longtable}
\usepackage{pifont}
\usepackage{soul}            % \hl highlighting
\usepackage{fontawesome5}    % icons
\IfFileExists{fontawesome5.sty}{
  \usepackage{fontawesome5} % gives \faIcon[<style>]{<name>}
}{
}

\definecolor{algoPurple}{HTML}{6A51A3}
\definecolor{algoBlue}{HTML}{1F77B4}
\definecolor{algoGreen}{HTML}{2E8B57}
\definecolor{algoOrange}{HTML}{E67E22}

\makeatletter
\@ifundefined{faBalanceScale}{
  \@ifundefined{faCalculator}{
    \@ifundefined{faChartLine}{
    }{}
  }{}
}{}

\@ifundefined{faBrain}{
  \@ifundefined{faLightbulb}{
    \@ifundefined{faCompass}{
      
    }{}
  }{}
}{}
\@ifundefined{faUsersCog}{
  \@ifundefined{faUsers}{
    \@ifundefined{faProjectDiagram}{
      
    }{}
  }{}
}{}

\@ifundefined{faThumbsUp}{
  \@ifundefined{faHandshake}{
    \@ifundefined{faCheckDouble}{
      
    }{}
  }{}
}{}
\makeatother

\definecolor{AbsBack}{HTML}{EEF2FF}   % light indigo-ish
\definecolor{AbsFrame}{HTML}{5A67D8}  % indigo frame
\definecolor{AbsTitle}{HTML}{3B49B1}  % darker title

\newtcolorbox{abstractbox}{
  enhanced, breakable,
  colback=AbsBack, colframe=AbsFrame!85,
  boxrule=0.7pt,
  borderline={0.5pt}{0pt}{AbsFrame!40},
  arc=8pt, left=10pt, right=10pt, top=10pt, bottom=2pt,
  drop fuzzy shadow=AbsFrame!25
}

\newcommand{\AbstractTitle}{\textbf{\textcolor{AbsTitle}{\fontsize{18}{18}\selectfont Abstract}}}

\usepackage{iftex}
\ifPDFTeX
  \PackageError{pragya-fonts}{You must compile with XeLaTeX or LuaLaTeX}{}
\fi

\usepackage{fontspec}
\defaultfontfeatures{Ligatures=TeX, Scale=MatchLowercase}

\newfontfamily\PragyaHeadline[
  Path=fonts/,
  UprightFont = PragyaHeadline-Regular.ttf,
  BoldFont    = PragyaHeadline-Bold.ttf
]{Pragya Headline}

\usepackage{titlesec}

\titleformat{\section}
  {\PragyaHeadline\bfseries\Large\color{AbsTitle}}
  {\textcolor{AbsTitle}{\thesection}}
  {0.6em}
  {}

\titleformat{\subsection}
  {\PragyaHeadline\bfseries\large\color{AbsTitle}}
  {\textcolor{AbsTitle}{\thesubsection}}
  {0.5em}
  {}

\titleformat{\subsubsection}
  {\PragyaHeadline\bfseries\normalsize\color{AbsTitle}}
  {\textcolor{AbsTitle}{\thesubsubsection}}
  {0.5em}
  {}

\titlespacing*{\section}{0pt}{1.0ex plus .2ex}{0.6ex}
\titlespacing*{\subsection}{0pt}{0.8ex plus .2ex}{0.4ex}
\titlespacing*{\subsubsection}{0pt}{0.6ex plus .1ex}{0.3ex}

\usepackage{empheq}              % for boxed equations
\usepackage[normalem]{ulem}      % for \ul

\usepackage{amsthm}        % provides \newtheorem and proof environment

\usepackage{fontspec}
\usepackage{polyglossia}

\setdefaultlanguage{english}
\setotherlanguage{hindi}
\newfontfamily\devanagarifont[
  Script=Devanagari,
  Path=./fonts/,
  UprightFont=NotoSerifDevanagari-Regular,
  BoldFont=NotoSerifDevanagari-Bold,
]{NotoSerifDevanagari-Regular}

\title{\textcolor{white}{.}}

\begin{document}
%\setcitestyle{square}
%\maketitle
\begin{figure*}[t]
  \centering
  \includegraphics[width=.98\linewidth]{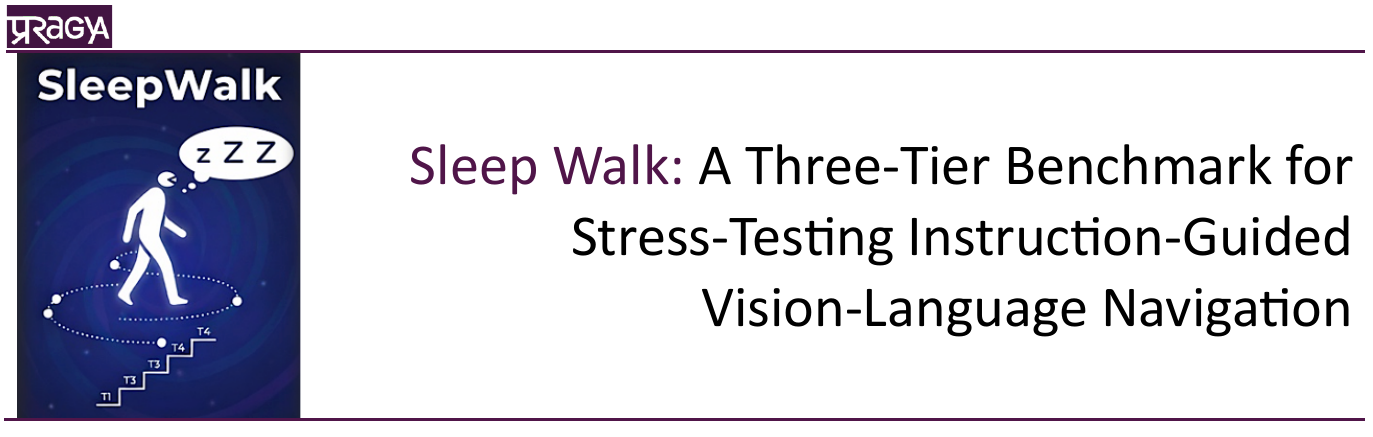}
  \vspace{-1.5em}
\end{figure*}

\begin{center}
{\Large\bfseries
Niyati Rawal\textsuperscript{\textcolor{blue}{$\dagger\star$}},
Sushant Ravva\textsuperscript{\textcolor{blue}{$\star$}},
Shah Alam Abir\textsuperscript{\textcolor{blue}{$\P$}},
Saksham Jain\textsuperscript{\textcolor{blue}{*}},
Aman Chadha\textsuperscript{\textcolor{blue}{$\ddagger$}}\footnote{Work done outside Apple, USA},
Vinija Jain\textsuperscript{\textcolor{blue}{$\S$}}\footnote{Work done outside Meta, USA},
Suranjana Trivedy\textsuperscript{\textcolor{blue}{$\star$}},
Amitava Das\textsuperscript{\textcolor{blue}{$\star$}}
\par}

{\Large
\textsuperscript{\textcolor{blue}{$\dagger$}}
Indian AI Research Organization (IAIRO), India,

\textsuperscript{\textcolor{blue}{$\star$}}
{\devanagarifont प्र}ragya Lab, BITS Pilani Goa, India,

\textsuperscript{\textcolor{blue}{$\P$}}
University of Dhaka, Bangladesh,

\textsuperscript{\textcolor{blue}{*}}
Delhi Technological University, India,

\textsuperscript{\textcolor{blue}{$\ddagger$}}
Apple, USA,

\textsuperscript{\textcolor{blue}{$\S$}}
Meta, USA
}
%\\
%\texttt{amitavad@goa.bits-pilani.ac.in}
\end{center}
\vspace{-1em} % small nudge only; avoid large negatives

\begin{abstractbox}
  \AbstractTitle
  \vspace{0.6em}
  \begin{spacing}{0.7}
  Vision-Language Models (VLMs) have advanced rapidly in multimodal perception and language understanding, yet it remains unclear whether they can reliably ground language into spatially coherent, plausibly executable actions in 3D digital environments. We introduce \textbf{\emph{SleepWalk}}, a benchmark for evaluating \textbf{\emph{instruction-grounded trajectory prediction}} in \textbf{\emph{single-scene 3D worlds}} generated from textual scene descriptions and filtered for navigability. Unlike prior navigation benchmarks centered on long-range exploration across rooms, SleepWalk targets \textbf{\emph{localized, interaction-centric embodied reasoning}}: given rendered visual observations and a natural-language instruction, a model must predict a trajectory that respects scene geometry, avoids collisions, and terminates at an action-compatible location. The benchmark covers diverse indoor and outdoor environments and organizes tasks into \textbf{\emph{three tiers of spatial and temporal difficulty}}, enabling fine-grained analysis of grounding under increasing compositional complexity. Using a standardized \textbf{\emph{pointwise judge-based evaluation protocol}}, we evaluate three frontier VLMs on \textbf{\emph{2,472 curated 3D environments}} with \textbf{\emph{nine instructions per scene}}. Results reveal \textbf{\emph{systematic failures in grounded spatial reasoning}}, especially under occlusion, interaction constraints, and multi-step instructions: performance drops as the difficulty level of the tasks increase.  In general, current VLMs can somewhat produce trajectories that are simultaneously spatially coherent, plausibly executable, and aligned with intended actions. By exposing failures in a controlled yet scalable setting, SleepWalk provides a critical benchmark for advancing grounded multimodal reasoning, embodied planning, vision-language navigation, and action-capable agents in 3D environments. \href{https://spatialreason123-prog.github.io/Sleepwalk-Reasoning-3DWorld/}{\textbf{\emph{Repository}}}
  \end{spacing}

%We propose a \textbf{game-theory–inspired} vulnerability metric with a formal \textbf{training$\rightarrow$deployment} asymmetry, and extend the equilibrium definition to admit both \textbf{Nash} and \textbf{Bayesian--Stackelberg} formulations; this yields a \textbf{closed-form threshold certificate} characterizing when models flip from refusal to unsafe compliance under \emph{training-simulation} cues. \textbf{Empirically,} we observe that \textbf{DPO}-style methods reduce cue-driven flipping relative to vanilla \textbf{SFT/BCO}, \textbf{KPO/KTO} further dampens cue sensitivity when kernels encode context invariance, while \textbf{GRPO} exhibits a bimodal pattern depending on whether rewards include \textbf{context-agnostic safety penalties}. \textbf{Larger models} are \emph{more} sensitive to training cues (higher posterior shift), and \textbf{Mixture-of-Experts (MoE)} variants tend to \textbf{amplify} train-simulation effects unless equipped with authenticated oversight or deception penalties. Overall, our results indicate that \textbf{authenticated training signals}, \textbf{penalizing unsafe compliance regardless of purported context}, and \textbf{explicit invariance to unauthenticated ``training'' cues} are critical to driving \textbf{FRACTURE\_FAKING} toward zero.
\end{abstractbox}

\section{Introduction}
\label{sec:intro}

\textbf{\emph{The next frontier for Vision-Language Models (VLMs) is not merely to describe the world, but to act in it with spatial precision.}} Recent progress in multimodal learning has substantially expanded the capabilities of VLMs in image captioning, visual question answering, and instruction following~\citep{vinyals2015show, antol2015vqa, anderson2018vision}. Yet as these models are increasingly positioned as the reasoning core of embodied agents, robotic assistants, and action-capable foundation models, a more consequential question comes into focus: \textbf{\emph{can they reliably translate natural-language instructions into spatially grounded, executable behavior in 3D environments?}}

\begin{figure}[t]
\centering
\includegraphics[width=\linewidth]{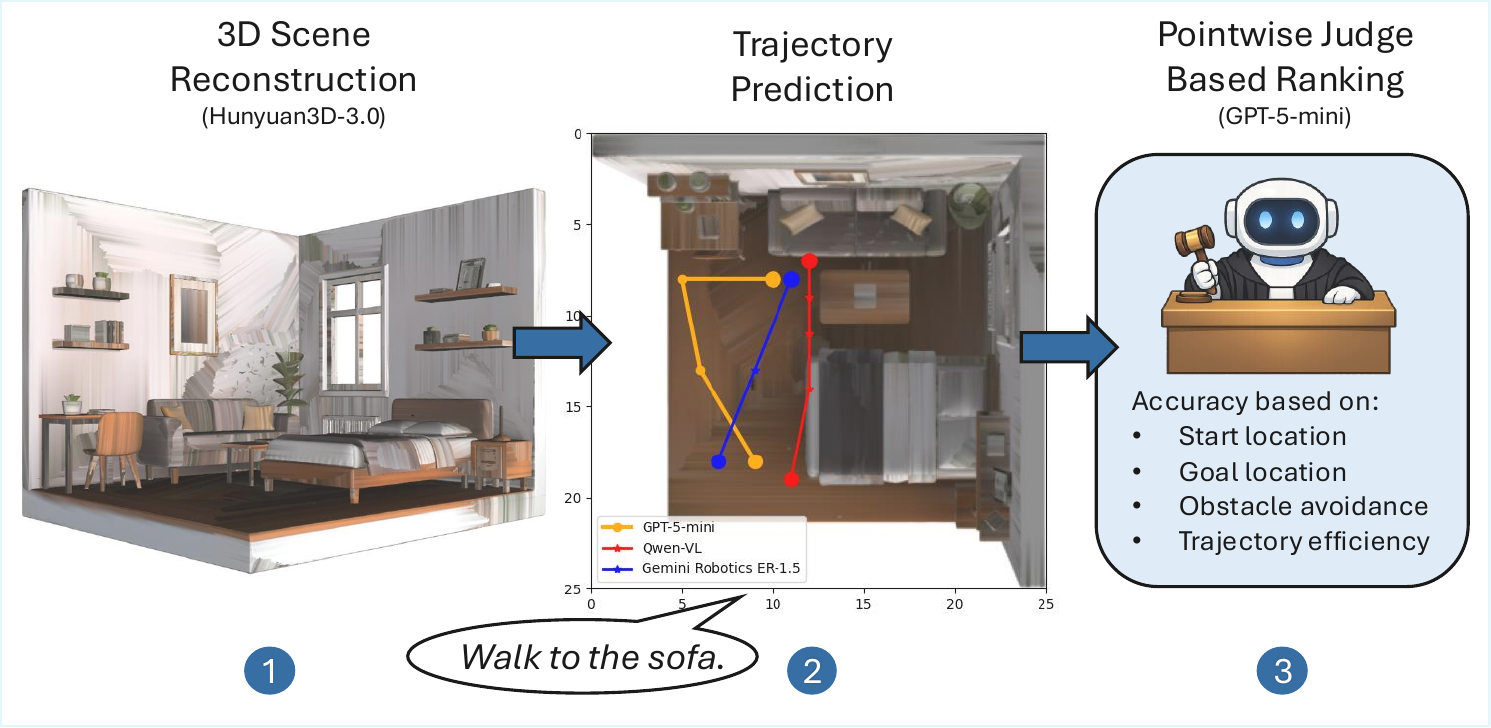}
\caption{\textit{Overview of SleepWalk.} A language instruction is converted into a single-scene 3D environment using Hunyuan3D-3.0. For each scene, given language instructions, different VLMs predict trajectories, which are visualized using top-down views. A fixed judge model scores trajectories in a pointwise manner, and rankings are aggregated across environments.}
\label{fig:overview}
\vspace{-1em}
\end{figure}

This problem is important because embodied competence demands far more than semantic recognition. An agent must localize itself, infer reachable goals, reason about geometry and occlusion, respect environmental constraints, and generate action sequences that remain feasible for downstream interaction. In other words, success depends on whether a model can connect \textbf{\emph{language, space, and action}} rather than merely align text with pixels. Recent work has shown that even strong VLMs continue to struggle with embodied spatial understanding, top-view reasoning, fine-grained navigation competence, multi-view robotic reasoning, and precise embodied grounding~\citep{du-etal-2024-embspatial,li-etal-2024-topviewrs,wang2024navigating,feng2025seeing,xue2025point}. These limitations matter directly for future systems that must plan, navigate, manipulate, and interact safely in the physical world.

A particularly important setting is \textbf{\emph{3D scene navigation under natural-language instructions}}. Classical Vision-and-Language Navigation (VLN) benchmarks have played a foundational role in embodied AI by studying how agents follow instructions in simulated environments~\citep{anderson2018vision}. However, much of this literature emphasizes long-horizon movement across rooms or buildings, where evaluation is often dominated by endpoint success or goal reachability. While such benchmarks remain essential, they provide only partial visibility into a harder and increasingly practical capability: \textbf{\emph{localized, interaction-centric reasoning within a single scene.}} In many real deployments, an embodied agent must do more than reach a destination. It must approach an object from a feasible direction, stop at an interaction-compatible location, avoid clutter and collisions, and preserve the geometric conditions necessary for the requested action.

Recent benchmarks have begun to expose this broader gap. EmbSpatial-Bench highlights weaknesses in embodied spatial understanding from egocentric 3D scenes~\citep{du-etal-2024-embspatial}; TopViewRS studies top-view reasoning relevant to localization and map-based navigation~\citep{li-etal-2024-topviewrs}; Navigating the Nuances shows that standard VLN evaluation can obscure systematic deficits in directional, landmark, and numerical instruction following~\citep{wang2024navigating}; and newer embodied benchmarks such as MV-RoboBench and Point-It-Out reveal persistent failures in multi-view robotic reasoning and staged visual grounding~\citep{feng2025seeing, xue2025point}. \textbf{\emph{What remains missing, however, is a benchmark that directly tests whether a model can produce a continuous, spatially coherent, executable trajectory inside a single 3D scene under object interaction and motion constraints.}}

To address this gap, we introduce \textbf{\emph{SleepWalk}}, a benchmark for evaluating \textbf{\emph{instruction-grounded trajectory prediction}} in \textbf{\emph{single-scene 3D environments}} reconstructed from language descriptions (Fig.~\ref{fig:overview}). Each environment corresponds to one coherent indoor or outdoor scene, deliberately avoiding room-to-room exploration in favor of \textbf{\emph{fine-grained spatial understanding, object-centric interaction, and trajectory feasibility}}. Given rendered visual observations and a natural-language instruction, a model must infer a continuous path that is consistent with scene layout, avoids collisions, and terminates at a location compatible with the intended action, such as approaching an object, picking it up, or sitting on furniture. Unlike symbolic planning or endpoint-only evaluation, SleepWalk assesses the \textbf{\emph{full spatial and temporal coherence}} of the predicted path with respect to both the environment and the instruction.

SleepWalk comprises \textbf{\emph{2,472 curated 3D environments}} spanning diverse layouts, object configurations, and levels of clutter. For each scene, we generate \textbf{\emph{nine instructions}} across \textbf{\emph{three tiers of difficulty}}, enabling controlled analysis of embodied reasoning under increasing compositional and interaction complexity. We further visualize selected trajectories with MotionGPT and TLControl to inspect whether predicted paths remain compatible with humanoid execution~\citep{jiang2024motiongpt,wan2024tlcontrol}. To compare heterogeneous model outputs, we introduce a standardized judge-based evaluation protocol that scores trajectories by start-location consistency, goal satisfaction, obstacle avoidance, and trajectory efficiency. Across frontier VLMs, we observe substantial degradation under \textbf{\emph{occlusion, interaction constraints, and multi-step instructions}}, even when the same models perform reasonably on simple goal-reaching prompts.

\textbf{\emph{The broader significance of SleepWalk is that it operationalizes a critical missing layer between seeing and acting.}} It is not only a benchmark for navigation, but a diagnostic testbed for future research on \textbf{\emph{grounded multimodal reasoning, embodied planning, robot instruction following, action-aware world modeling, and spatially reliable multimodal agents.}} As VLMs move from passive perception toward embodied deployment, the ability to evaluate this intermediate layer---where language must be converted into feasible behavior---will become increasingly central to both capability and safety.

The contributions of this paper are threefold:
\begin{itemize}
    \item We introduce \textbf{\emph{SleepWalk}}, a new benchmark that constructs spatially consistent \textbf{\emph{single-scene 3D environments}} from language descriptions using Hunyuan3D-3.0, and generates diverse trajectory-following instructions using Qwen3-8B-VL, yielding \textbf{\emph{nine instructions per environment}} for multi-level embodied evaluation.
    \item We benchmark frontier \textbf{\emph{Vision-Language Models}} on \textbf{\emph{continuous coordinate-based trajectory prediction}} from visual observations and natural-language instructions, emphasizing \textbf{\emph{spatial feasibility, temporal coherence, and interaction compatibility}} across easy, medium, and hard tasks.
    \item We introduce a standardized \textbf{\emph{judge-based evaluation protocol}} for comparing heterogeneous model outputs, and use it to reveal persistent limitations in \textbf{\emph{grounded spatial reasoning}}, especially in \textbf{\emph{occluded, compositional, and interaction-heavy}} scenarios.
\end{itemize}
\section{SleepWalk: A Three-Tier Benchmark for Grounded Trajectory Reasoning}
\label{sec:method}

\begin{figure}[t]
\centering
\includegraphics[width=\linewidth]{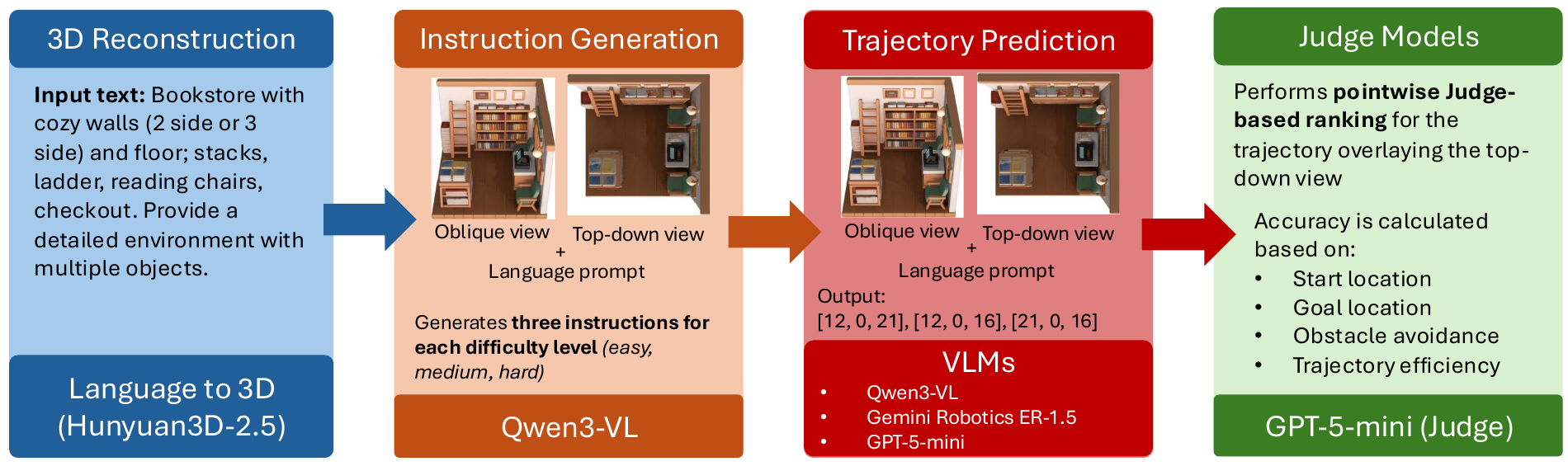}
\caption{\textbf{\emph{SleepWalk pipeline.}} Starting from a natural-language scene description, we reconstruct a single-scene 3D environment with Hunyuan3D-3.0, render top-down and oblique observations, and use Qwen3-8B-VL to generate tiered navigation instructions (\emph{easy}, \emph{medium}, \emph{hard}). Given the scene views and an instruction, a VLM predicts a continuous action trajectory, which is then evaluated by a judge model (GPT-5-mini) using pointwise scoring over \emph{start-location consistency}, \emph{goal satisfaction}, \emph{obstacle avoidance}, and \emph{trajectory efficiency}.}
\label{fig:method}
\vspace{-1em}
\end{figure}

\textbf{\emph{SleepWalk is a benchmark for testing whether Vision-Language Models can convert language into spatially grounded, executable behavior in 3D environments.}} As illustrated in Fig.~\ref{fig:method}, the benchmark begins from natural-language scene descriptions, reconstructs navigable single-scene 3D worlds, generates tiered navigation instructions, and evaluates predicted trajectories under a standardized protocol. The design of SleepWalk deliberately emphasizes \textbf{\emph{localized, interaction-centric reasoning}} rather than long-range exploration: a model must understand scene geometry, identify feasible routes, avoid obstacles, and stop at a location compatible with the requested action. In this way, SleepWalk probes a critical capability that lies between \textbf{\emph{seeing}} and \textbf{\emph{acting}}. This focus is motivated by a growing body of recent evidence showing that current VLMs still struggle with embodied spatial grounding, top-view reasoning, and fine-grained navigation competence, even when they perform strongly on broader multimodal tasks~\citep{du-etal-2024-embspatial,li-etal-2024-topviewrs,wang2024navigating,mayer2025ivispar}.

\subsection{Text-to-3D Environment Reconstruction}

Each SleepWalk instance begins with a natural-language scene description. We sample MS-COCO captions and manually filter or rewrite them to obtain $1{,}200$ descriptions suitable for \textbf{\emph{navigable single-scene 3D generation}}, spanning both indoor and outdoor settings. We then use Hunyuan3D-3.0~\citep{cao2025hunyuanimage} to convert each description into a spatially coherent 3D environment by estimating scene layout, object placement, and overall geometry.

The resulting environments are intentionally restricted to \textbf{\emph{single coherent scenes}} rather than multi-room worlds. This keeps the benchmark focused on the regime of greatest interest: fine-grained movement, local planning, and action feasibility around objects, furniture, and clutter. By combining indoor and outdoor scenes, SleepWalk captures a broader range of layouts, densities, and navigation constraints, thereby increasing both environmental diversity and reasoning difficulty. This single-scene emphasis complements prior benchmarks that evaluate embodied spatial understanding from egocentric views~\citep{du-etal-2024-embspatial} or top-view map reasoning~\citep{li-etal-2024-topviewrs}, while moving closer to continuous action prediction in executable 3D scenes.

\subsection{Instruction Generation}

For each reconstructed scene, we generate navigation instructions using Qwen3-8B-VL~\citep{yang2025qwen3}. The model is given both \textbf{\emph{top-down}} and \textbf{\emph{oblique}} views of the environment and is prompted to produce \textbf{\emph{nine goal-directed instructions}} organized into three difficulty tiers: \emph{easy}, \emph{medium}, and \emph{hard}.

The generated instructions are designed to remain \textbf{\emph{scene-grounded, concise, and executable}}. They target behaviors such as approaching an object, moving between landmarks, manipulating an item, or reaching a location compatible with sitting or interaction. Difficulty increases with compositional and spatial demands: easy instructions typically require short-range goal localization, medium instructions involve structured spatial dependencies, and hard instructions introduce multi-step goals, interaction constraints, or longer planning horizons. This three-tier design enables controlled analysis of how model performance degrades as embodied reasoning becomes more demanding, in the same spirit as recent fine-grained evaluations that diagnose failure modes beyond coarse task success~\citep{wang2024navigating,mayer2025ivispar}.

\subsection{Action-Conditioned Trajectory Prediction}

Given a reconstructed 3D environment $\mathcal{E}$ and a natural-language instruction $\mathcal{I}$, the task is to predict a continuous trajectory
\[
\mathcal{T} = \{p_t\}_{t=1}^{T}, \qquad p_t \in \mathbb{R}^{3},
\]
where $p_t$ denotes the agent position at time step $t$. A valid trajectory must satisfy three conditions: it must be \textbf{\emph{spatially feasible}} within $\mathcal{E}$, it must \textbf{\emph{avoid collisions}} with scene elements, and it must terminate at a location that supports execution of the intended action described by $\mathcal{I}$.

The model receives rendered visual observations $\mathcal{V}$ from the environment together with the instruction, and predicts
\[
\mathcal{T} = f_{\theta}(\mathcal{V}, \mathcal{I}).
\]
All evaluated VLMs are tested in a \textbf{\emph{frozen, zero-shot setting}}: model parameters $\theta$ remain fixed, and no task-specific fine-tuning or adaptation is performed on SleepWalk.

This formulation is intentionally stricter than endpoint-based navigation. A model is not rewarded merely for ending near the correct goal; instead, it must produce a trajectory whose \textbf{\emph{entire path}} is consistent with scene geometry, object affordances, and temporal ordering. SleepWalk therefore evaluates whether a VLM can ground language into a \textbf{\emph{full sequence of spatially and temporally coherent actions}}, rather than only a final destination. Relative to recent benchmarks that study embodied spatial understanding~\citep{du-etal-2024-embspatial}, top-view reasoning~\citep{li-etal-2024-topviewrs}, or fine-grained VLN diagnostics~\citep{wang2024navigating}, SleepWalk directly targets the missing intermediate layer between instruction interpretation and executable path generation.

\subsection{Judge-Based Evaluation Protocol}

To assess whether predicted trajectories are grounded, executable, and instruction-consistent, we introduce a standardized \textbf{\emph{judge-based evaluation protocol}}. Rather than relying only on heuristic metrics such as distance to goal, we use a strong vision-language model as a structured evaluator that jointly considers the instruction, the scene, and the predicted path.

For each candidate trajectory, the judge is given:  
(i) the \textbf{\emph{top-down view}} of the reconstructed 3D environment,  
(ii) the \textbf{\emph{natural-language instruction}}, and  
(iii) the \textbf{\emph{trajectory overlay} rendered on the map}.

Each trajectory is represented as an ordered sequence of 3D waypoints projected onto the 2D top-down plane. The judge is prompted to evaluate whether the trajectory aligns with the instruction while respecting scene geometry and environmental constraints.

We use a \textbf{\emph{pointwise scoring scheme}} in which each trajectory is evaluated independently. The judge assigns a score for four factors:
\begin{itemize}
    \item \textbf{\emph{Start Location Consistency}}: Does the trajectory begin at the correct initial region?
    \item \textbf{\emph{Goal Satisfaction}}: Does the trajectory end at a location that satisfies the instruction?
    \item \textbf{\emph{Obstacle Avoidance}}: Does the path avoid collisions and remain geometrically plausible?
    \item \textbf{\emph{Trajectory Efficiency}}: Is the route reasonably direct, without unnecessary detours?
\end{itemize}

Each factor is explicitly defined in the evaluation prompt to reduce ambiguity and improve reproducibility. For a trajectory $\tau$, the judge assigns a discrete score
\begin{equation*}
s_k(\tau) \in \{1,2,3,4,5\} \cup \{\mathrm{N/A}\},
\qquad
k \in \{\text{start}, \text{goal}, \text{obs}, \text{eff}\}.
\end{equation*}

Valid scores are normalized to the interval $[0,1]$:
\begin{equation*}
\tilde{s}_k(\tau) = \frac{s_k(\tau)}{5}.
\end{equation*}

Let $\mathcal{T}_{t,k}^{\mathrm{valid}}$ denote the set of trajectories in difficulty tier $t$ with valid (non-\text{N/A}) scores for factor $k$. The tier-level factor score is then
\begin{equation*}
S_k(t)=\frac{1}{|\mathcal{T}_{t,k}^{\mathrm{valid}}|}
\sum_{\tau \in \mathcal{T}_{t,k}^{\mathrm{valid}}} \tilde{s}_k(\tau).
\end{equation*}

To summarize performance across difficulty tiers, we compute the overall factor score
\begin{equation*}
S_k=\frac{1}{|\mathcal{T}_{k}^{\mathrm{tiers}}|}
\sum_{t \in \mathcal{T}_{k}^{\mathrm{tiers}}} S_k(t),
\end{equation*}
where $\mathcal{T}_{k}^{\mathrm{tiers}} \subseteq \{\text{easy}, \text{medium}, \text{hard}\}$ denotes the set of tiers with valid evaluations for factor $k$. By default, all tiers are weighted equally.

\textbf{\emph{This protocol allows heterogeneous trajectory outputs to be compared under a shared notion of grounded success.}} Crucially, it evaluates not only where a model ends up, but whether it reaches that endpoint through a path that is spatially plausible and action-compatible. This is especially important in light of recent work showing that coarse task-level metrics can mask systematic spatial and navigational failures~\citep{wang2024navigating,mayer2025ivispar}.

\subsection{Humanoid Visualization and Embodied Execution}

To assess whether predicted trajectories are compatible with embodied execution, we visualize selected outputs using a humanoid pipeline based on TLControl~\citep{wan2024tlcontrol} and MotionGPT~\citep{jiang2024motiongpt}. TLControl converts waypoint sequences into control signals, while MotionGPT generates realistic full-body motions such as walking and object interaction.

This stage is used for \textbf{\emph{qualitative validation}} rather than primary scoring. Its purpose is to reveal whether trajectories that appear reasonable in top-down space remain plausible when executed by an embodied humanoid. In practice, this visualization helps expose failures that may be less obvious in static overlays, such as collisions with nearby objects, awkward stopping positions, or motion patterns incompatible with the intended interaction.

\textbf{\emph{Overall, SleepWalk provides a controlled yet scalable benchmark for grounded trajectory reasoning in Vision-Language Models}}, combining 3D reconstruction, tiered instructions, trajectory prediction, structured evaluation, and embodied visualization to study the translation of language into \textbf{\emph{feasible spatial behavior}}.

% To further inspect whether predicted trajectories are compatible with embodied execution, we visualize selected trajectories using a humanoid control and animation pipeline built on TLControl\citep{wan2024tlcontrol} and MotionGPT\citep{jiang2024motiongpt}. Given a predicted waypoint sequence, TLControl converts the trajectory into low-level control signals for navigation and interaction, while MotionGPT synthesizes realistic full-body motion for actions such as walking, reaching, picking up objects, or sitting.

% This stage is used for \textbf{\emph{qualitative validation}} rather than primary scoring. Its purpose is to reveal whether trajectories that appear reasonable in top-down space remain plausible when executed by an embodied humanoid. In practice, this visualization helps expose failures that may be less obvious in static overlays, such as collisions with nearby objects, awkward stopping positions, or motion patterns incompatible with the intended interaction.

% \textbf{\emph{Taken together, these components make SleepWalk a controlled yet scalable benchmark for studying grounded trajectory reasoning in Vision-Language Models.}} By combining single-scene 3D reconstruction, tiered instruction generation, continuous path prediction, structured evaluation, and embodied visualization, SleepWalk isolates a capability that is becoming increasingly central to future multimodal systems: the ability to transform language into \textbf{\emph{feasible spatial behavior}}.
%\input{2_attack_taxonomy}

\section{Evaluating Grounded Trajectory Reasoning in 3D Scenes}
\label{sec:experiments}

\textbf{\emph{We now ask a central empirical question: can current Vision-Language Models translate instructions into trajectories that are spatially grounded, physically feasible, and compatible with the intended action in 3D scenes?}} To answer this, we evaluate three representative frontier VLMs---\textbf{\emph{Qwen3-VL}}, \textbf{\emph{Gemini Robotics ER-1.5}}, and \textbf{\emph{GPT-5-mini}}---on SleepWalk.

Because SleepWalk is designed as an evaluation benchmark rather than a training resource, all experiments are conducted in a strictly \textbf{\emph{evaluation-only, zero-shot setting}}. No model is fine-tuned or adapted to the benchmark. For each scene--instruction pair, each model produces a single trajectory under \textbf{\emph{deterministic decoding}}, ensuring that comparisons remain reproducible and attributable to capability rather than sampling variance. Each predicted trajectory is rendered using a standardized visual format consisting of one top-down view and one oblique scene view, preserving both geometric structure and task context.

We use \textbf{\emph{GPT-5-mini}} as the judge model with a fixed evaluation prompt %and temperature set to $0$ 
throughout. The judge scores each predicted trajectory independently using the pointwise evaluation protocol defined in Section~\ref{sec:method}. All evaluated models are tested under identical conditions---the same reconstructed scenes, the same instructions, the same trajectory representation, and the same judge configuration---so that observed differences reflect \textbf{\emph{differences in grounded trajectory reasoning rather than experimental mismatch}}.

\begin{figure}[t]
\centering

\begin{tabular}{cccc}
\subcaptionbox{}{\includegraphics[width=0.20\textwidth]{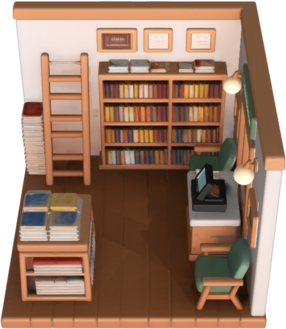}} &
\subcaptionbox{}{\includegraphics[width=0.20\textwidth]{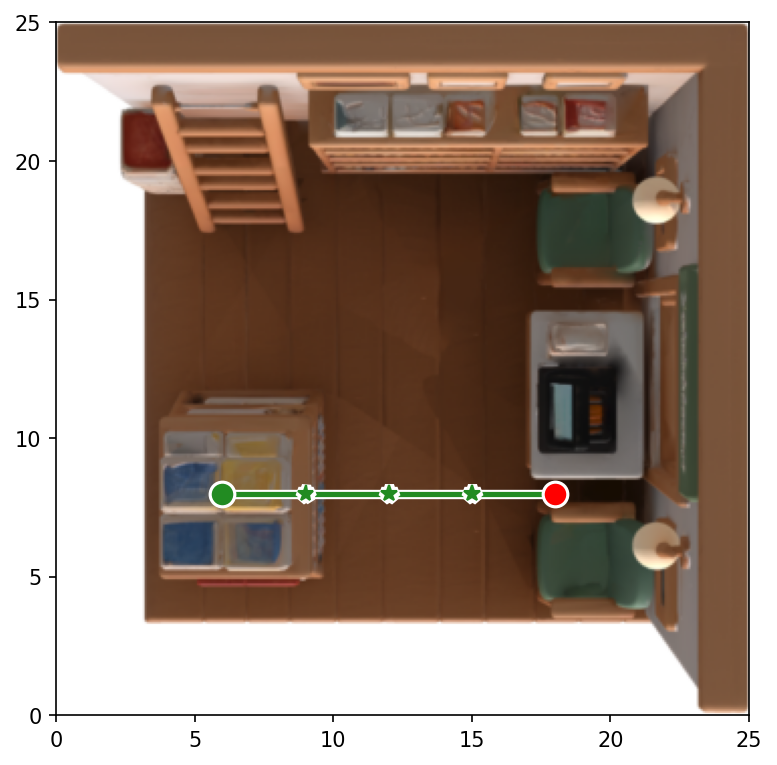}} &
\subcaptionbox{}{\includegraphics[width=0.20\textwidth]{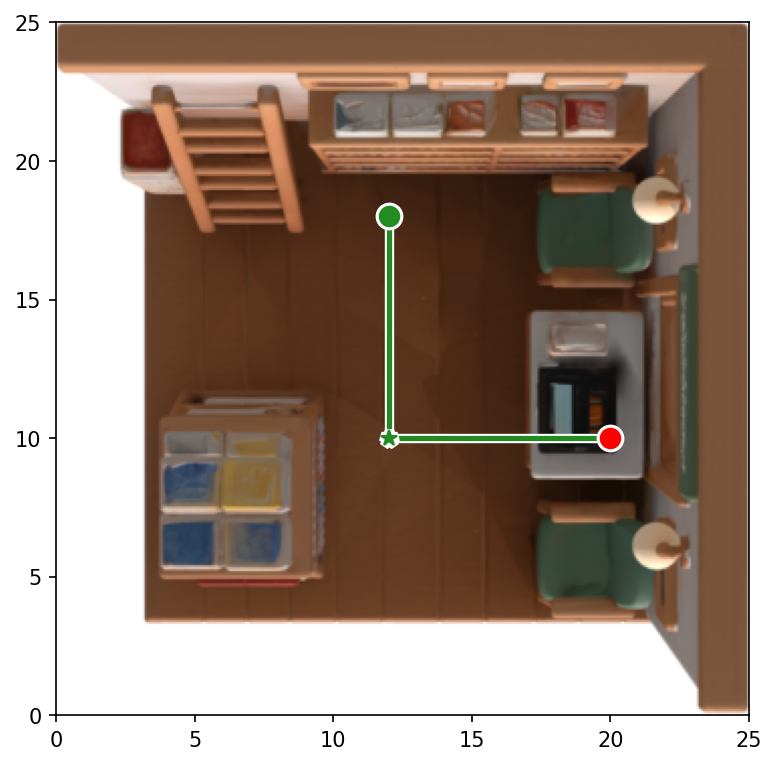}} &
\subcaptionbox{}{\includegraphics[width=0.20\textwidth]{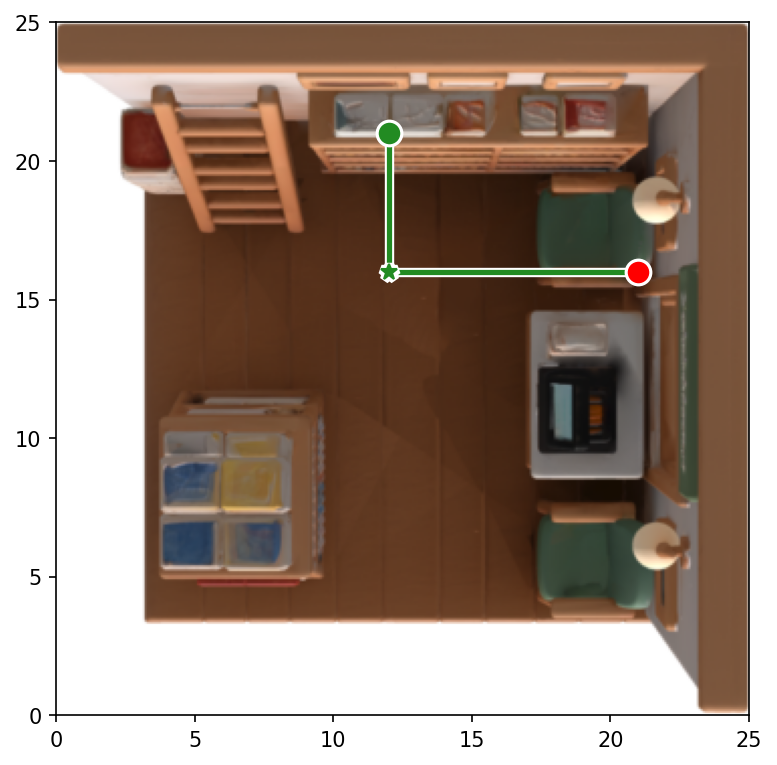}} \\
\end{tabular}

\caption*{\textit{LEVEL 1 $|$ TASK: Walk from the bookshelf to the wall-mounted lamp $|$ START: Near the bookshelf $|$ END: Near the wall-mounted lamp $|$ INTERACT: none}}

\setcounter{subfigure}{0}
\begin{tabular}{cccc}
\subcaptionbox{}{\includegraphics[width=0.20\textwidth]{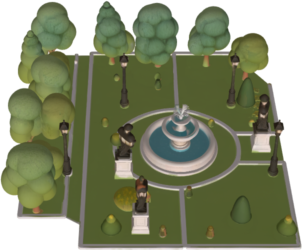}} &
\subcaptionbox{}{\includegraphics[width=0.20\textwidth]{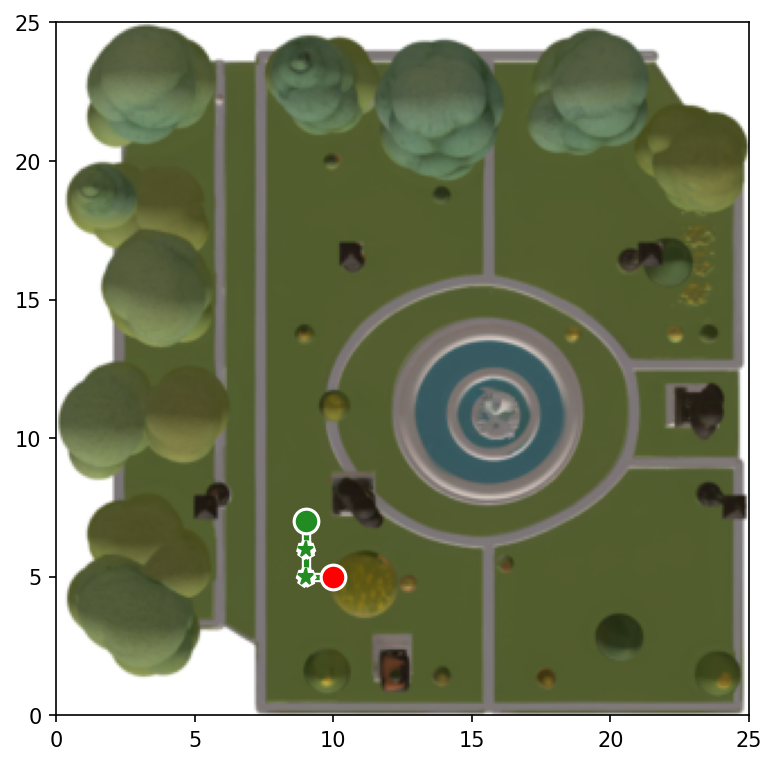}} &
\subcaptionbox{}{\includegraphics[width=0.20\textwidth]{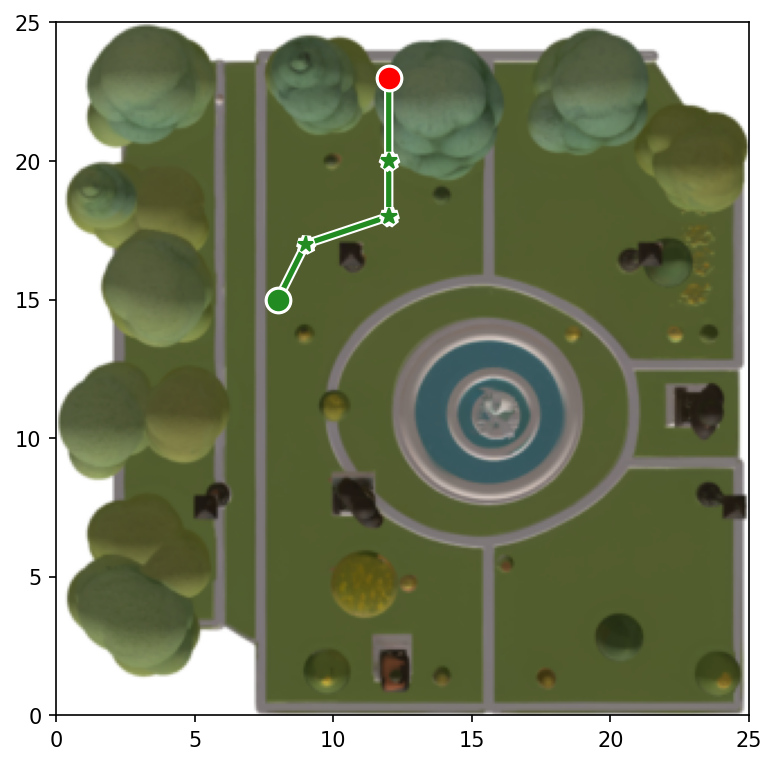}} &
\subcaptionbox{}{\includegraphics[width=0.20\textwidth]{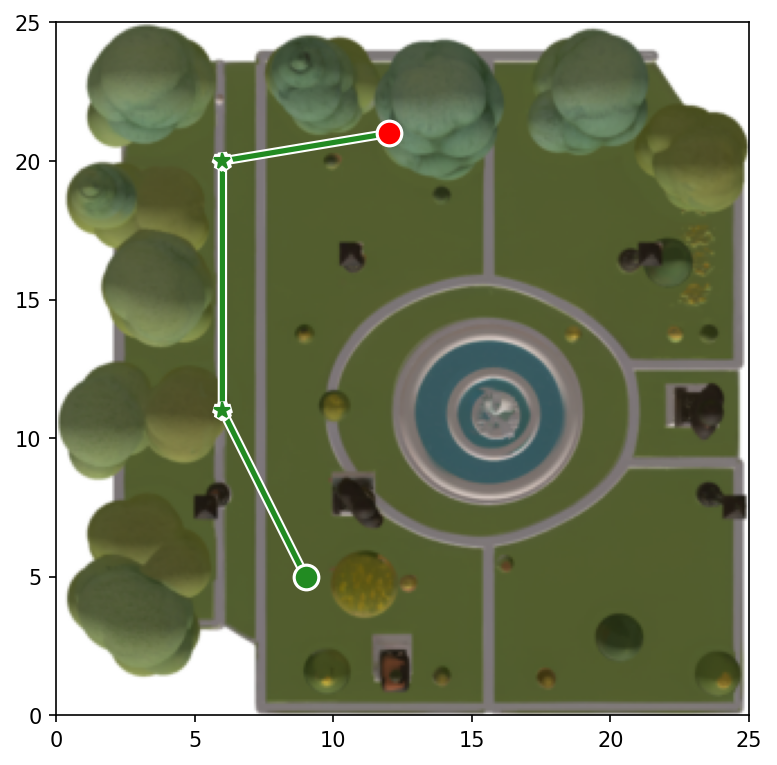}} \\
\end{tabular}
\caption*{\textit{LEVEL 2 $|$ TASK: Approach the yellow spherical object and then move to the northern tree $|$ START: Near the yellow spherical object $|$ END: Near the northern tree $|$ INTERACT: yellow spherical object, tree}}

\setcounter{subfigure}{0}
\begin{tabular}{cccc}
\subcaptionbox{}{\includegraphics[width=0.20\textwidth]{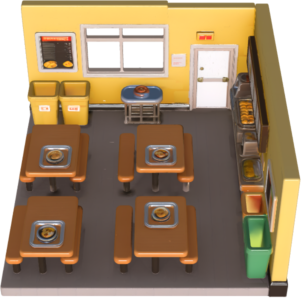}} &
\subcaptionbox{}{\includegraphics[width=0.20\textwidth]{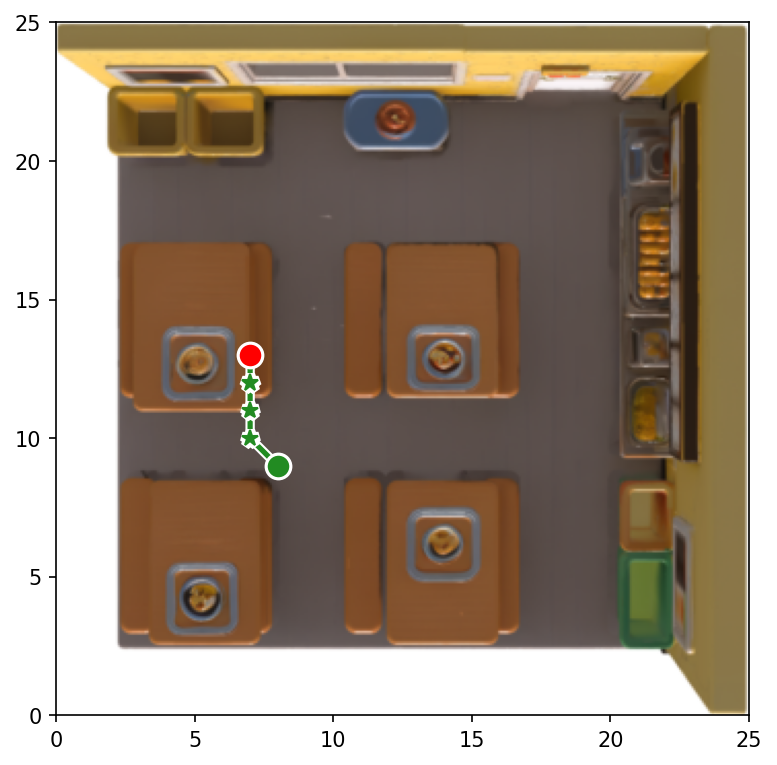}} &
\subcaptionbox{}{\includegraphics[width=0.20\textwidth]{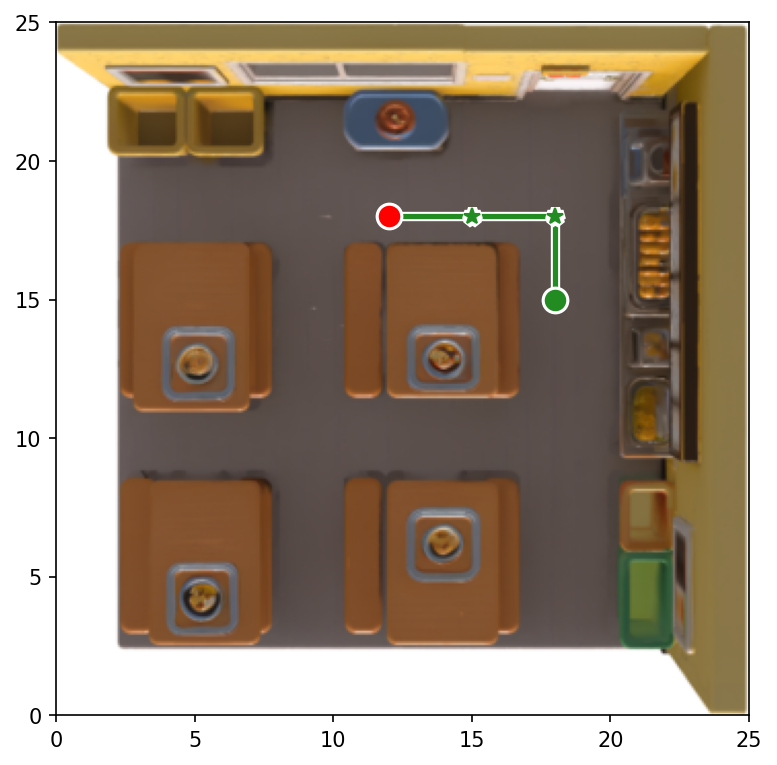}} &
\subcaptionbox{}{\includegraphics[width=0.20\textwidth]{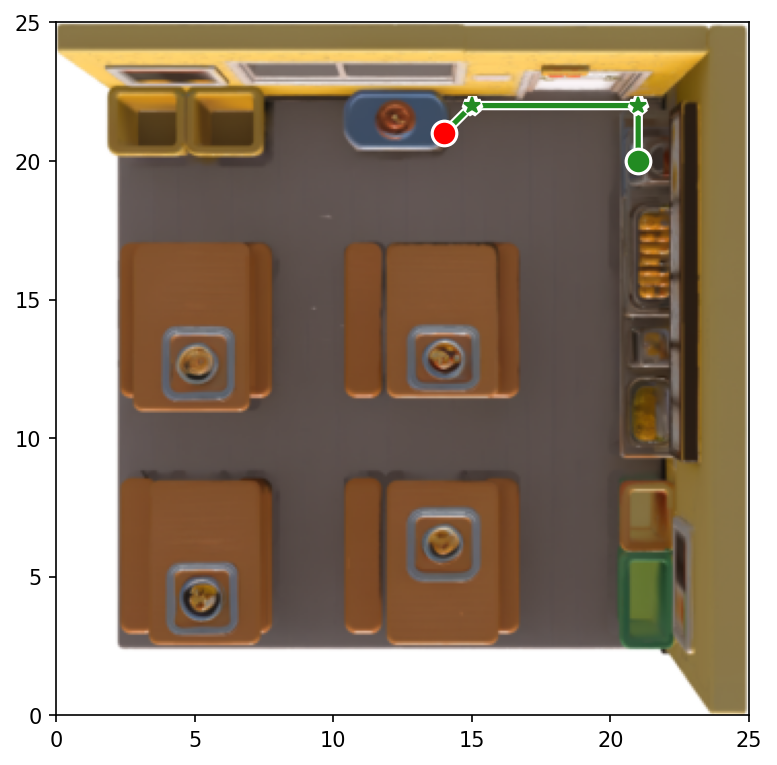}} \\
\end{tabular}
\caption*{\textit{LEVEL 3 $|$ TASK: Pick up the tray from the service counter, walk to the round table, and place it there $|$ START: Near the service counter $|$ END: Near the round table $|$ INTERACT: tray, service counter, round table}}

\caption{\textbf{\emph{Qualitative trajectory comparison across difficulty levels.}} Column (a) shows the reconstructed scene view, while columns (b), (c), and (d) show trajectory predictions from Qwen3-VL, Gemini Robotics ER-1.5, and GPT-5-mini, respectively. Across levels, the figure highlights differences in start-location grounding, goal accuracy, obstacle avoidance, and robustness under compositional instructions.}
\label{fig:qual}
\vspace{-1em}
\end{figure}

\subsection{Qualitative Results}

\textbf{\emph{Figure~\ref{fig:qual} shows a clear pattern: as task complexity increases, failures shift from coarse recognition to grounded execution.}} We analyze one example from each difficulty tier to illustrate how different VLMs translate instructions into trajectories.

For the \textbf{\emph{easy}} example, \textit{``Walk from the bookshelf to the wall-mounted lamp,''} all three models recover the broad task semantics: they localize the bookshelf region and produce plausible routes toward a lamp. GPT-5-mini selects the more plausible target lamp, whereas Gemini Robotics ER-1.5 ends closer to the farther lamp. However, \textbf{\emph{none of the models fully satisfies local physical constraints}}: parts of their trajectories lie on or too close to the source or target object, which would induce collision during execution. Even simple tasks therefore reveal a gap between \textbf{\emph{semantic grounding}} and \textbf{\emph{physically feasible motion}}.

The \textbf{\emph{medium}} example, \textit{``Approach the yellow spherical object and then move to the northern tree,''} requires resolving two grounded references while preserving temporal order. Qwen3-VL fails on both ends, while Gemini Robotics ER-1.5 identifies the final tree more accurately but misses the correct start region. GPT-5-mini is the only model that captures both. Here, the dominant failure is not collision, but \textbf{\emph{compositional grounding}}: models struggle to bind intermediate references, preserve order, and assign correct spatial roles.

The \textbf{\emph{hard}} example, \textit{``Pick up the tray from the service counter, walk to the round table, and place it there,''} introduces explicit interaction and multi-step planning. GPT-5-mini again produces the strongest trajectory, correctly aligning both start and goal while maintaining a plausible route. \textbf{\emph{Across tiers, the trend is consistent: as instructions become more interaction-heavy and temporally structured, errors compound rapidly.}}

Taken together, these examples expose three recurring failure modes in SleepWalk: \textbf{\emph{mislocalized starts, incomplete or incorrect goal grounding, and trajectories that appear semantically plausible yet remain physically unsafe.}} The qualitative evidence therefore supports the main claim of this benchmark: \textbf{\emph{current VLMs often understand what an instruction refers to, but still struggle to convert that understanding into spatially coherent, executable behavior in 3D scenes.}}

\subsection{Overall Model Ranking}

Table~\ref{tab:overall_ranking} reports overall factor scores for Qwen3-VL, Gemini Robotics ER-1.5, and GPT-5-mini. Scores lie in $[0,1]$, with higher values indicating better performance. \textbf{\emph{Among the models with complete quantitative results, GPT-5-mini performs best on all four factors}}: start-location consistency, goal satisfaction, obstacle avoidance, and trajectory efficiency.

The pattern is also diagnostic. Both models score relatively high on obstacle avoidance, but much lower on goal grounding. \textbf{\emph{This suggests that producing a navigable-looking path is easier than reaching the correct interaction-compatible endpoint.}} Thus, the main bottleneck is not coarse geometry, but \textbf{\emph{precise instruction grounding.}}

\begin{table}[t]
\centering
\caption{\textbf{\emph{Model Comparison on SleepWalk.}} Higher scores indicate better performance.}
\label{tab:overall_ranking}
\begin{tabular}{lcccc}
\toprule
\textbf{Model} & \textbf{Start loc.} & \textbf{Goal loc.} & \textbf{Avoid obs.} & \textbf{Traj. eff.}\\
\midrule
Qwen3-VL & 0.48 & 0.20 & 0.84 & 0.47 \\
Gemini Robotics ER-1.5 & 0.58 & 0.34 & 0.89 & 0.58 \\
GPT-5-mini & 0.75 & 0.51 & 0.91 & 0.64 \\
\bottomrule
\end{tabular}
\end{table}

\begin{figure}[t]
\centering
\includegraphics[width=\linewidth]{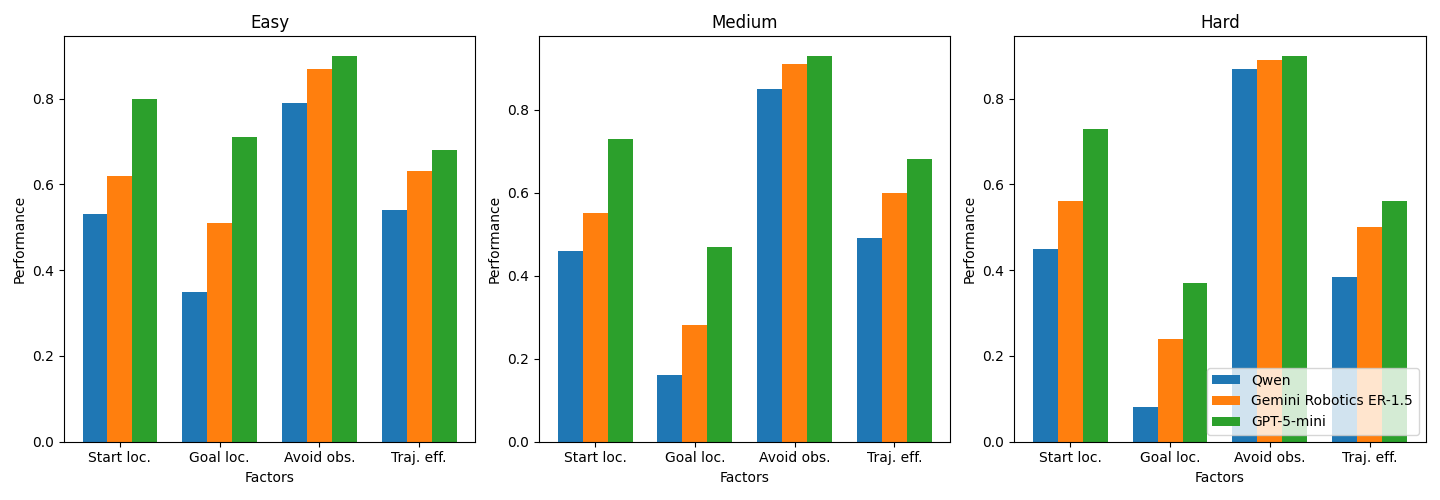}
\caption{\textbf{\emph{Factor-wise performance across difficulty tiers.}} Comparison of Qwen3-VL, Gemini Robotics ER-1.5, and GPT-5-mini on four evaluation factors---start location, goal location, obstacle avoidance, and trajectory efficiency---across easy, medium, and hard tiers.}
\label{fig:quant}
\vspace{-1em}
\end{figure}

\subsection{Ranking by Difficulty Level}

Figure~\ref{fig:quant} breaks down performance by difficulty tier across the four evaluation factors. \textbf{\emph{The trend is clear: performance drops from easy to medium to hard tasks.}} This indicates that SleepWalk successfully exposes increasing demands on grounded spatial reasoning.

Among the quantitatively reported models, \textbf{\emph{GPT-5-mini performs best across factors and tiers}}, while Qwen3-VL degrades more sharply, especially on goal grounding and trajectory efficiency. The steepest drop occurs on harder tasks requiring multi-step reasoning, ordered execution, and interaction-aware endpoint selection.

Taken together, the quantitative results reinforce the qualitative findings: current VLMs retain competence on simple goal-reaching behavior, but deteriorate sharply once tasks require \textbf{\emph{compositional grounding, interaction awareness, and temporally structured planning}}.

\subsection{From Predicted Paths to Embodied Execution}

% \begin{figure}[t]
% \centering
% \includegraphics[width=0.4\linewidth]{figures/colm/visualize.png}
% \caption{\textbf{\emph{From predicted path to humanoid execution.}} We animate the trajectory generated by GPT-5-mini for the instruction \textit{``Walk from the bookshelf to the wall-mounted lamp''} using TLControl~\citep{wan2024tlcontrol} and MotionGPT~\citep{jiang2024motiongpt}. This provides a qualitative check of whether a path that appears correct in top-down space remains plausible when executed by a humanoid agent.}
% \label{fig:vis}
% \vspace{-1em}
% \end{figure}

\begin{figure}[t]
\centering

\begin{tabular}{cccc}
\subcaptionbox{}{\includegraphics[width=0.45\textwidth]{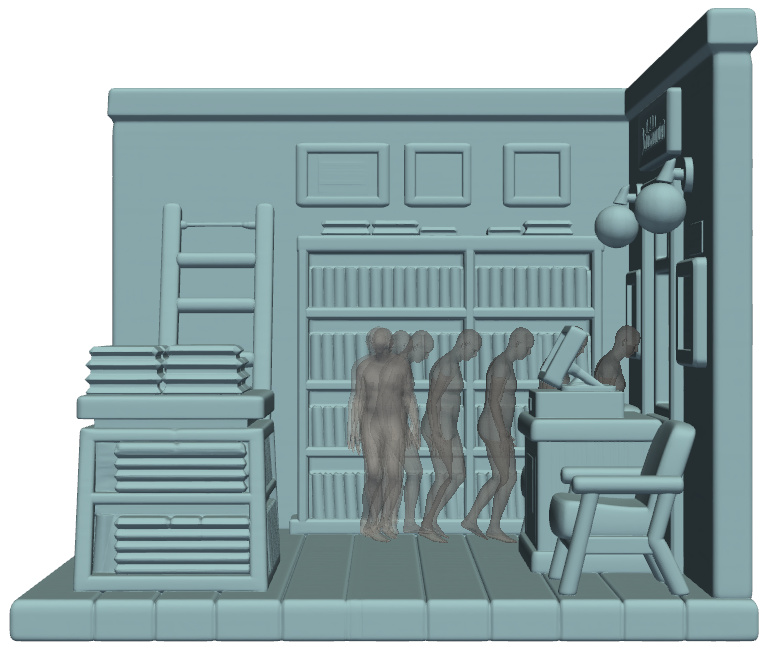}} &
\subcaptionbox{}{\includegraphics[width=0.45\textwidth]{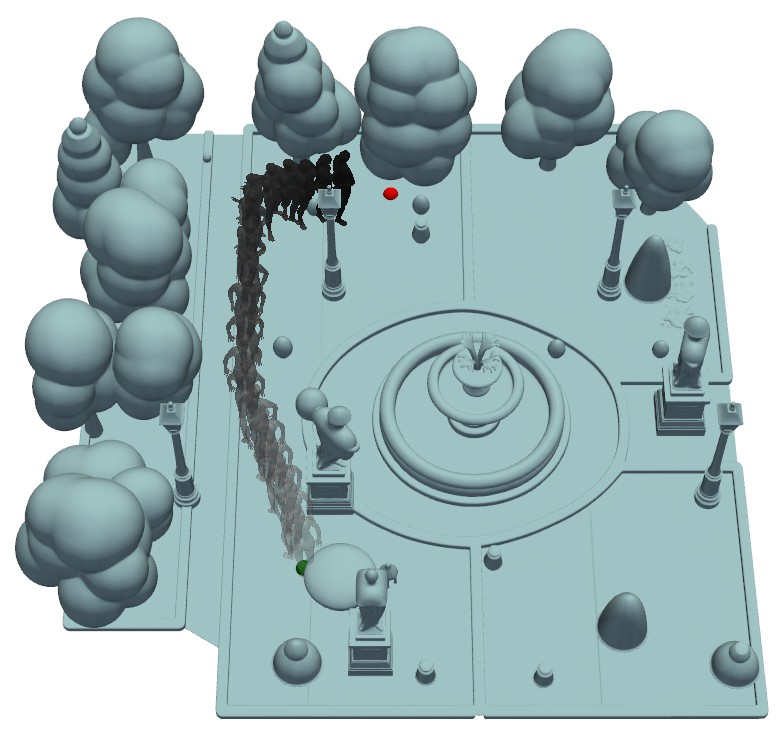}} 
\end{tabular}
\caption{\textbf{\emph{From predicted path to humanoid execution.}} We animate the trajectory generated by GPT-5-mini for the instruction \textit{``Walk from the bookshelf to the wall-mounted lamp''} and \textit{``Approach the yellow spherical object and then move to the northern tree''} using TLControl~\citep{wan2024tlcontrol} and MotionGPT~\citep{jiang2024motiongpt}. This provides a qualitative check of whether a path that appears correct in top-down space remains plausible when executed by a humanoid agent.}
\label{fig:vis}
\end{figure}

To move beyond static trajectory overlays, we perform a final \textbf{\emph{embodied executability check}} using a humanoid control and animation pipeline. Specifically, we take the waypoint sequence predicted by GPT-5-mini, project it back into the reconstructed 3D environment, and render it as humanoid motion. TLControl~\citep{wan2024tlcontrol} converts the trajectory into low-level control signals, while MotionGPT~\citep{jiang2024motiongpt} synthesizes realistic full-body movement conditioned on the predicted path.

Figure~\ref{fig:vis} shows two such examples for the instruction \textit{``Walk from the bookshelf to the wall-mounted lamp.''} and \textit{``Approach the yellow spherical object and then move to the northern tree''}. The resulting animation allows us to inspect whether trajectories that appears reasonable in top-down space remains \textbf{\emph{physically plausible}} when executed by an embodied agent. This matters because geometrically valid paths can still fail at execution time due to collisions, awkward stopping positions, or unnatural motion transitions.

\textbf{\emph{This stage provides a qualitative bridge between geometric correctness and embodied feasibility.}} It does not replace the benchmark’s primary judge-based evaluation, but serves as an additional validation layer for whether predicted paths can support realistic action in 3D environments. In practice, it makes failure modes more visible, especially in cases where a trajectory is semantically plausible yet operationally unsafe or physically incompatible with the intended interaction.

\section{Conclusion and Future Avenues}

\textbf{\emph{SleepWalk benchmarks whether VLMs can turn language into grounded, executable trajectories in 3D scenes.}} By focusing on \textbf{\emph{localized, interaction-centric reasoning}}, it isolates a key gap between perception and action. Across models, we find that current VLMs still struggle with \textbf{\emph{spatial grounding, compositional instructions, and executable path generation}}, especially in occluded and multi-step settings.

Promising next steps include \textbf{\emph{(i)}} richer multi-view and temporal observations, \textbf{\emph{(ii)}} direct reasoning over structured 3D scene representations, \textbf{\emph{(iii)}} tighter coupling between predicted paths and downstream motion or control models, and \textbf{\emph{(iv)}} extending the benchmark into physics simulators for embodied training and sim-to-real transfer.

\textbf{\emph{Overall, SleepWalk provides a scalable testbed for advancing grounded multimodal reasoning, embodied planning, and action-capable vision-language systems.}}
%\input{5_alkali_dataset}
%\input{6_experiments}
%\input{7_dpo_aqi}
%\input{7_results}
%\input{8_conclusion}

%\newpage
%\input{9_discussion}

%\subsection{\textcolor{orange}{Objective Manipulation}}
%----------------------

\newpage
\bibliographystyle{acl_natbib}
\bibliography{biblio}

\clearpage
\newpage

\section{Frequently Asked Questions (FAQs)}
\label{sec:FAQs}

\begin{itemize}

% ===================== Q1 =====================
\item[$\blacktriangleright$] \textbf{What is the main scientific contribution of the paper? The current presentation combines scene generation, instruction generation, trajectory prediction, judge-based scoring, and empirical analysis, making it difficult to tell whether the contribution is the benchmark, the pipeline, or the findings.}
\begin{description}
\item[\ding{224}] \emph{Short answer.}
The \textbf{primary contribution} is the \textbf{\emph{benchmark formulation itself}}: a controlled evaluation setting for testing whether Vision--Language Models can convert natural-language instructions into \textbf{\emph{continuous, spatially grounded, and plausibly executable trajectories}} in \textbf{\emph{single-scene 3D environments}}. The generation pipeline instantiates this benchmark at scale, and the empirical study demonstrates the failure modes it reveals. \hfill \emph{[Supported by benchmark motivation and stated contributions]} 

\emph{Clarification.}
More precisely, the paper contributes:
\begin{enumerate}
    \item a \textbf{\emph{single-scene, interaction-centric benchmark setting}} for grounded trajectory reasoning,
    \item a \textbf{\emph{scalable construction pipeline}} that reconstructs 3D scenes, generates tiered instructions, and evaluates predicted paths, and
    \item an \textbf{\emph{empirical diagnosis}} showing that current frontier VLMs degrade substantially on compositional, interaction-heavy, and multi-step instructions.
\end{enumerate}

\emph{Why this matters.}
The paper should not be read as claiming that any one component---for example, the judge or the scene generator---is the sole novelty. Rather, the novelty lies in defining and instantiating a benchmark that measures a capability that existing evaluation protocols often miss: whether a model can map language into \textbf{\emph{locally executable spatial behavior}} rather than merely recognizing semantics or reaching a rough endpoint.

\emph{Takeaway.}
The paper is best understood first as a \textbf{\emph{benchmark for grounded trajectory reasoning in 3D scenes}}, with the pipeline and experiments serving that central benchmark contribution.
\end{description}

% ===================== Q2 =====================
\item[$\blacktriangleright$] \textbf{How is SleepWalk meaningfully different from existing Vision-and-Language Navigation benchmarks? At first glance, this appears to be another instruction-following navigation task in 3D environments.}
\begin{description}
\item[\ding{224}] \emph{Short answer.}
The key distinction is that SleepWalk targets \textbf{\emph{localized, interaction-centric reasoning within a single scene}}, whereas much prior VLN work emphasizes \textbf{\emph{long-horizon movement across rooms or buildings}} and often evaluates success primarily through endpoint reachability. SleepWalk instead requires that the \textbf{\emph{entire trajectory}} be spatially coherent, obstacle-aware, and compatible with the intended action. \hfill \emph{[Supported by Introduction]} 

\emph{Why this is not just a smaller VLN benchmark.}
The central challenge here is not global exploration, but whether the model can:
\begin{itemize}
    \item identify the correct local target,
    \item approach it through a feasible route,
    \item avoid clutter and collisions, and
    \item stop at a location that is compatible with the requested action.
\end{itemize}
This is precisely the regime where coarse endpoint-only evaluation can hide important failures.

\emph{Why this matters.}
Many embodied tasks depend on this intermediate capability between passive perception and full action execution. A system may know \emph{what} object is relevant, yet still fail to plan a path that reaches it correctly, safely, and in the right temporal order.

\emph{Takeaway.}
The benchmark is not merely a narrower VLN instance; it is designed to probe a \textbf{\emph{different layer of embodied competence}}: converting language into \textbf{\emph{continuous, executable local motion}}. 
\end{description}

% ===================== Q3 =====================
\item[$\blacktriangleright$] \textbf{Why restrict the environments to a single coherent scene? Doesn’t that make the problem less realistic than full multi-room or building-scale navigation?}
\begin{description}
\item[\ding{224}] \emph{Short answer.}
The \textbf{single-scene restriction is deliberate}. The paper is not trying to replace long-horizon navigation benchmarks; it is isolating a complementary capability: \textbf{\emph{fine-grained local reasoning around objects, obstacles, and interaction targets}}. \hfill \emph{[Supported by Section 2.1]} 

\emph{Why this is a principled design choice.}
Restricting the task to a single scene reduces confounds from:
\begin{itemize}
    \item room-to-room exploration,
    \item large-scale search,
    \item map-building over long horizons, and
    \item global navigation heuristics that can obscure local failures.
\end{itemize}
That makes it easier to study whether a model can actually produce a path that is \textbf{\emph{interaction-compatible}} at the local level.

\emph{Why this matters.}
A large fraction of real embodied failures occur not because the system cannot traverse a building, but because it cannot correctly execute the \emph{last few meters}: approaching the wrong side of an object, stopping too close to clutter, colliding with nearby items, or violating temporal structure in a multi-step task.

\emph{Takeaway.}
The single-scene focus should be read as \textbf{\emph{diagnostic sharpening}}, not as simplification for its own sake. It allows the benchmark to study the gap between \textbf{\emph{seeing}} and \textbf{\emph{acting}} in a controlled way. 
\end{description}

% ===================== Q4 =====================
\item[$\blacktriangleright$] \textbf{The benchmark is highly synthetic: scenes are generated from text, instructions are generated by a model, and evaluation is also model-based. How can we be confident that the paper is measuring grounded reasoning rather than artifacts of the generation pipeline?}
\begin{description}
\item[\ding{224}] \emph{Short answer.}
The paper’s claim is \textbf{not} that the synthetic pipeline is artifact-free or that it fully substitutes for real-world embodied evaluation. The narrower claim is that synthetic construction provides a \textbf{\emph{controlled and scalable way}} to build a diagnostic benchmark for a capability that is currently under-measured: translating language into feasible spatial behavior inside 3D scenes. \hfill \emph{[Supported by pipeline description and benchmark framing]} 

\emph{Why the synthetic pipeline is used.}
The design supports:
\begin{itemize}
    \item scalable benchmark construction,
    \item controlled task generation,
    \item diverse layouts and clutter conditions, and
    \item systematic difficulty tiering across easy, medium, and hard tasks.
\end{itemize}

\emph{Why the claim remains meaningful.}
The scientific question here is not whether the scenes are indistinguishable from all real environments. It is whether they are coherent enough to test whether a model can:
\begin{itemize}
    \item interpret an instruction,
    \item reason over rendered scene observations,
    \item generate a collision-aware path, and
    \item terminate at an action-compatible location.
\end{itemize}
That is a benchmark question, and a synthetic but curated pipeline can still be a valid way to study it.

\emph{Takeaway.}
The paper should be read as presenting a \textbf{\emph{controlled diagnostic testbed}}, not as claiming to exhaust all forms of embodied evaluation. 
\end{description}

% ===================== Q5 =====================
\item[$\blacktriangleright$] \textbf{Why use a judge model at all? Wouldn’t automatic metrics such as endpoint distance, collision count, or shortest-path deviation be more objective and reproducible than GPT-5-mini scoring trajectories?}
\begin{description}
\item[\ding{224}] \emph{Short answer.}
Automatic geometric metrics are useful, but they are \textbf{\emph{insufficient on their own}} for the capability this benchmark targets. A trajectory may be short, collision-free, and geometrically plausible while still failing the instruction because it reaches the wrong object, ignores a required intermediate subgoal, or ends at a location from which the intended action cannot plausibly be executed. The judge is therefore introduced to capture \textbf{\emph{instruction consistency and interaction compatibility}} in addition to geometry. \hfill \emph{[Supported by evaluation protocol]} 

\emph{Why the judge is used here.}
The paper explicitly scores four factors:
\begin{enumerate}
    \item \textbf{\emph{Start Location Consistency}},
    \item \textbf{\emph{Goal Satisfaction}},
    \item \textbf{\emph{Obstacle Avoidance}}, and
    \item \textbf{\emph{Trajectory Efficiency}}.
\end{enumerate}
This means the evaluation is designed to reflect not only where a model ends up, but whether it gets there through a path that is spatially plausible and aligned with the intended action.

\emph{What the paper does and does not claim.}
The manuscript does \emph{not} claim that judge-based evaluation is perfect. It makes the narrower methodological choice of using a \textbf{\emph{fixed judge and fixed prompt %and fixed temperature
}} to obtain a standardized scoring protocol for heterogeneous path outputs. 

\emph{Takeaway.}
The judge is used not because geometric metrics are unimportant, but because \textbf{\emph{grounded path correctness in this task is partly semantic and action-dependent}}, not purely geometric.
\end{description}

% ===================== Q6 =====================
\item[$\blacktriangleright$] \textbf{A central concern is trustworthiness of the judge. Why should readers trust GPT-5-mini as evaluator, especially when the main empirical conclusions depend on its judgments?}
\begin{description}
\item[\ding{224}] \emph{Short answer.}
The paper’s position is limited and operational: GPT-5-mini is used as a \textbf{\emph{standardized evaluation instrument}} under fixed conditions, not as an infallible ground-truth oracle. The core argument is that, for this task, a structured judge is a practical way to compare heterogeneous trajectories under a shared notion of grounded success. \hfill \emph{[Supported by Section 2.4 and Section 3 setup]} 

\emph{What is actually claimed in the paper.}
The paper states that:
\begin{itemize}
    \item GPT-5-mini is used as the judge model,
    \item the evaluation prompt is fixed, and
    %\item the temperature is set to 0, and
    \item all models are evaluated under identical conditions.
\end{itemize}
This supports consistency of comparison.

\emph{What the paper does not currently claim.}
The visible manuscript text does not provide human-agreement studies, multi-judge comparisons, or prompt-sensitivity analyses. Accordingly, the rebuttal should not overstate validation that is not in the paper.

\emph{Takeaway.}
The judge should be understood as a \textbf{\emph{practical standardized evaluator for a difficult multimodal task}}, not as proof that every score is final or uniquely correct.
\end{description}

% ===================== Q7 =====================
\item[$\blacktriangleright$] \textbf{Could poor performance simply reflect an output-interface problem? Since models are asked to emit explicit coordinate trajectories, perhaps they understand the task but fail because waypoint prediction is a difficult formatting requirement.}
\begin{description}
\item[\ding{224}] \emph{Short answer.}
The benchmark intentionally evaluates \textbf{\emph{explicit trajectory prediction}} because the paper’s goal is to test whether models can map language and visual context into \textbf{\emph{concrete spatial behavior}}. In embodied systems, a trajectory is not merely a textual explanation of intent; it is the object that downstream execution would actually require. \hfill \emph{[Supported by task definition]} 

\emph{Why this is more than a formatting issue.}
The reported failures are not described as random syntax errors. Instead, the paper emphasizes substantive error patterns:
\begin{itemize}
    \item mislocalized start regions,
    \item incorrect or incomplete goal grounding,
    \item failure to preserve temporal order, and
    \item semantically plausible but physically unsafe trajectories.
\end{itemize}
These are grounding and planning failures, not merely output-format failures. 

\emph{Why the quantitative pattern matters.}
The finding that obstacle avoidance is comparatively stronger than goal grounding suggests a structured failure pattern: models can often generate \emph{some} navigable-looking path, but struggle to generate the \emph{correct} path for the intended action. 

\emph{Takeaway.}
While interface burden is a fair consideration, the evidence in the paper points to a deeper issue: \textbf{\emph{the main bottleneck is precise instruction-grounded spatial reasoning, not merely coordinate formatting}}.
\end{description}

% ===================== Q8 =====================
\item[$\blacktriangleright$] \textbf{How do we know that the easy/medium/hard instruction tiers are meaningful? Could these levels simply be heuristic labels rather than real differences in embodied reasoning difficulty?}
\begin{description}
\item[\ding{224}] \emph{Short answer.}
The tiers are motivated by increasing \textbf{\emph{compositional and spatial demand}}: easy tasks focus on short-range goal localization, medium tasks add structured spatial dependencies, and hard tasks introduce multi-step goals, interaction constraints, or longer planning horizons. \hfill \emph{[Supported by Section 2.2]} 

\emph{Why the tiering is empirically meaningful.}
The quantitative and qualitative results both support this organization:
\begin{itemize}
    \item performance drops from easy to medium to hard,
    \item harder tasks expose sharper degradation in goal grounding and trajectory efficiency, and
    \item qualitative examples show that the dominant failure modes become more compositional and interaction-heavy as difficulty increases.
\end{itemize}

\emph{Why this matters.}
The benchmark is valuable not only because models fail, but because the failure pattern degrades in a structured and interpretable way across increasing task demands.

\emph{Takeaway.}
The difficulty tiers are not merely cosmetic labels; they are a \textbf{\emph{diagnostic axis}} that helps reveal how grounded trajectory reasoning deteriorates as the reasoning burden grows.
\end{description}

% ===================== Q9 =====================
\item[$\blacktriangleright$] \textbf{Why are all experiments conducted in a frozen, zero-shot setting? Wouldn’t benchmark-specific fine-tuning or adaptation provide a more complete picture of what the task is measuring?}
\begin{description}
\item[\ding{224}] \emph{Short answer.}
The paper uses a \textbf{\emph{strict evaluation-only, zero-shot setting}} because its immediate goal is diagnostic: to measure whether current VLMs already possess this capability, not to study how well they can be adapted to it. \hfill \emph{[Supported by task setup]} 

\emph{Why this choice is methodologically useful.}
A zero-shot setup isolates pre-existing model capability and avoids conflating:
\begin{itemize}
    \item general grounded reasoning ability,
    \item adaptation to benchmark-specific distributions, and
    \item improvements driven by task-specific tuning.
\end{itemize}

\emph{Why this matters.}
Because SleepWalk is introduced first as a benchmark, it is sensible to establish that it reveals nontrivial failure modes even before any model is tuned to it. Training-time studies can naturally follow in later work.

\emph{Takeaway.}
The zero-shot setting is a \textbf{\emph{deliberate diagnostic choice}}, not an omission of a required training result.
\end{description}

% ===================== Q10 =====================
\item[$\blacktriangleright$] \textbf{Only three frontier models are evaluated. Is that enough to support the paper’s broader claims, or is the empirical evidence too narrow?}
\begin{description}
\item[\ding{224}] \emph{Short answer.}
For the current paper’s purpose, yes. The experiments are used to show that SleepWalk is \textbf{\emph{nontrivial, unsaturated, and capable of revealing structured failure modes}} even in strong contemporary systems. The paper does not claim to provide the final exhaustive leaderboard across all model families. \hfill \emph{[Supported by evaluation section]} 

\emph{Why this is sufficient for the paper’s scope.}
The empirical role of the model set is to establish that:
\begin{enumerate}
    \item the benchmark differentiates strong models,
    \item the revealed failures are systematic rather than trivial, and
    \item performance degrades under higher compositional and interaction demands.
\end{enumerate}

\emph{What the results already show.}
Among the evaluated models, GPT-5-mini performs best across all four reported factors, but even the strongest system degrades as the tasks become harder. This supports the benchmark’s central diagnostic claim. 

\emph{Takeaway.}
Broader model coverage would certainly strengthen future benchmarking, but the present set is sufficient to establish the \textbf{\emph{benchmark’s diagnostic value}}.
\end{description}

% ===================== Q11 =====================
\item[$\blacktriangleright$] \textbf{The paper reports that obstacle avoidance is relatively strong while goal grounding is much weaker. Why is that distinction important, and what does it tell us about current VLMs?}
\begin{description}
\item[\ding{224}] \emph{Short answer.}
This distinction is important because it separates \textbf{\emph{generic path plausibility}} from \textbf{\emph{instruction-faithful grounded behavior}}. The reported pattern suggests that current VLMs can often produce a route that looks navigable, yet still fail to reach the \textbf{\emph{correct interaction-compatible endpoint}}. \hfill \emph{[Supported by Section 3.2]} 

\emph{Why this matters scientifically.}
If models were failing mainly because they could not generate motion at all, then obstacle avoidance and route plausibility would also be weak. But the paper finds a more specific bottleneck: the hard part is not merely moving through space, but binding the instruction to the right object, preserving the right sequence of actions, and terminating at the right place.

\emph{What this implies.}
The main weakness exposed by SleepWalk is therefore not coarse navigation, but \textbf{\emph{precise grounded reasoning under action constraints}}.

\emph{Takeaway.}
This is one of the paper’s most informative findings: current VLMs are better at generating \emph{a plausible path} than generating \emph{the correct grounded path}. 
\end{description}

% ===================== Q12 =====================
\item[$\blacktriangleright$] \textbf{What role does the humanoid visualization stage actually play? Is it scientifically central, or is it mainly a demonstration?}
\begin{description}
\item[\ding{224}] \emph{Short answer.}
The humanoid stage is a \textbf{\emph{qualitative validation layer}}, not the primary evaluation protocol. Its purpose is to test whether trajectories that look reasonable in top-down space remain plausible when executed as humanoid motion. \hfill \emph{[Supported by Section 2.5 and Section 3.4]} 

\emph{Why it is included.}
This layer helps expose failures that may be less obvious in a static overlay, including:
\begin{itemize}
    \item near-collisions,
    \item awkward stopping positions,
    \item motion patterns incompatible with the intended interaction, and
    \item trajectories that are geometrically plausible yet operationally unsafe.
\end{itemize}

\emph{What the paper does not claim.}
The paper explicitly states that this stage is used for qualitative validation rather than for primary scoring.

\emph{Takeaway.}
Readers should interpret the humanoid visualization as \textbf{\emph{supporting qualitative evidence}} that strengthens the benchmark’s embodied interpretation, not as the benchmark’s central quantitative mechanism.
\end{description}

% ===================== Q13 =====================
\item[$\blacktriangleright$] \textbf{The dataset construction details are not fully transparent: the paper starts from 1{,}000 MS-COCO captions but later reports 2{,}472 curated 3D environments. Does this undermine confidence in the benchmark construction?}
\begin{description}
\item[\ding{224}] \emph{Short answer.}
It does \textbf{not} undermine the benchmark’s core idea, but it is a place where the presentation should be made clearer. The visible text indicates that the pipeline begins from \textbf{\emph{1{,}000 MS-COCO captions}} that are \textbf{\emph{manually filtered or rewritten}} for navigable single-scene generation, and later reports a final benchmark of \textbf{\emph{2{,}472 curated 3D environments}}. \hfill \emph{[Supported by Section 2.1 and earlier benchmark summary]} 

\emph{How to interpret this conservatively.}
The reasonable reading is that the construction process expands from an initial description pool into a larger curated environment set through generation and curation. However, the exact mapping should be documented more explicitly in the paper.

\emph{Why this matters.}
This is primarily a \textbf{\emph{transparency and documentation issue}}, not a conceptual flaw in the benchmark formulation itself.

\emph{Takeaway.}
The benchmark’s contribution remains intact, but the manuscript would benefit from a clearer description of the relationship between source descriptions, generation stages, filtering, and final retained environments.
\end{description}

% ===================== Q14 =====================
\item[$\blacktriangleright$] \textbf{What is the main empirical takeaway of the paper? Is the claim simply that current VLMs are weak at navigation, or something more specific?}
\begin{description}
\item[\ding{224}] \emph{Short answer.}
The paper makes a more specific claim than ``current VLMs are bad at navigation.'' Its core finding is that strong VLMs may understand the broad semantics of an instruction and even generate roughly navigable paths, yet still struggle to produce \textbf{\emph{precise, spatially grounded, interaction-compatible trajectories}} in 3D scenes, especially under \textbf{\emph{occlusion, compositional constraints, and multi-step planning demands}}. \hfill \emph{[Supported by Abstract and Results]} 

\emph{Why this is the right interpretation.}
The qualitative analysis identifies recurring failures in:
\begin{itemize}
    \item start grounding,
    \item goal grounding,
    \item temporal ordering, and
    \item physical executability.
\end{itemize}
The factor-wise quantitative results reinforce this by showing strong relative performance on obstacle avoidance but weaker performance on goal satisfaction. 

\emph{Takeaway.}
The central message is that there remains a measurable gap between \textbf{\emph{semantic understanding}} and \textbf{\emph{grounded executable behavior}}. SleepWalk is valuable because it makes that gap visible in a controlled and scalable way.
\end{description}

\end{itemize}
\clearpage
%\section{Appendix}
\appendix
% Reset appendix numbering in a standard conference style
\makeatletter
\@addtoreset{figure}{section}
\@addtoreset{table}{section}
\makeatother
\renewcommand{\thefigure}{\thesection\arabic{figure}}
\renewcommand{\thetable}{\thesection\arabic{table}}
\numberwithin{equation}{section}

\section{Appendix}
\label{app:construction}

This appendix provides additional details on benchmark construction, prompt design,
evaluation, and reproducibility. The goal is to make the setup of \benchmarkname{}
easy to understand, reproduce, and extend.

\subsection{Benchmark overview}
\label{app:overview}

\benchmarkname{} evaluates whether a vision-language model can convert a natural-language
instruction and rendered scene observations into a spatially coherent, executable trajectory
inside a single-scene 3D environment. The benchmark is intentionally designed to emphasize
localized, interaction-centric reasoning rather than long-range room-to-room exploration.

\begin{table}[htbp]
\centering
\small
\caption{Core benchmark statistics reported in the current paper.}
\label{tab:benchmark_stats}
\begin{tabular}{lc}
\toprule
Statistic & Value \\
\midrule
Number of environments & \numenvs \\
Difficulty tiers & 3 \\
Instructions per scene & \instrperscene \\
Instructions per tier & 3 \\
Total scene--instruction pairs & 22,248 \\
Rendered views per scene & 2 \\
Primary evaluation factors & 4 \\
\bottomrule
\end{tabular}
\end{table}

\subsection{Scene generation and curation}
\label{app:scene_generation}

Each benchmark instance begins from a textual scene description specifying a single coherent
indoor or outdoor environment. We convert the description into a 3D scene using
\scenegenmodel{}. Because the paper focuses on instruction-grounded local navigation and
interaction, we intentionally exclude multi-room or highly fragmented generations that would
shift the task toward global exploration rather than precise within-scene grounding.

After generation, scenes are manually filtered for the following properties:
\begin{itemize}
    \item \textbf{Single-scene coherence:} the environment should correspond to one visually and
    spatially coherent scene rather than multiple disconnected spaces.
    \item \textbf{Navigability:} the scene should contain sufficient free space for a humanoid-scale
    agent to move through the environment.
    \item \textbf{Object recognizability:} major landmarks and target objects should be visually
    identifiable from rendered views.
    \item \textbf{Geometric plausibility:} scenes with severe self-intersections, floating objects,
    degenerate layouts, or obviously broken geometry are removed.
\end{itemize}

For each accepted environment, we render two views: a top-down view and an oblique view.
The top-down view provides an explicit picture of free space, clutter, and approximate route
structure, while the oblique view provides appearance cues, object identity, and interaction
context.

\subsection{Instruction generation}
\label{app:instruction_generation}

For each reconstructed scene, we use \instrgenmodel{} to generate nine navigation
instructions, grouped into three difficulty tiers: easy, medium, and hard. The tiering is meant
to expose progressively harder forms of embodied reasoning rather than merely longer paths.

\begin{table}[htbp]
\centering
\small
\caption{Task-tier design in \benchmarkname{}.}
\label{tab:tier_definition}
\begin{tabular}{p{0.12\linewidth}p{0.26\linewidth}p{0.22\linewidth}p{0.28\linewidth}}
\toprule
Tier & Primary capability & Typical instruction type & Common failure mode exposed \\
\midrule
Easy & Single-goal localization & Move to a clearly identifiable object or region & Incorrect endpoint despite broadly plausible path \\
Medium & Compositional grounding & Resolve two landmarks or a simple ordered objective & Failure to preserve reference binding or temporal order \\
Hard & Multi-step interaction-aware planning & Pick/place, approach-then-move, or action-compatible stopping & Endpoint/action mismatch, unsafe stopping position, or incoherent route \\
\bottomrule
\end{tabular}
\end{table}

We instruct the generation model to produce concise, scene-grounded, and executable
instructions. Instructions are rejected if they refer to nonexistent objects, require ambiguous
global references, or describe actions that cannot be approximately assessed from the rendered
views.

\subsection{Trajectory representation}
\label{app:trajectory_representation}

Given rendered observations $V$ and a natural-language instruction $I$, a tested model predicts
a continuous trajectory
\begin{equation}
T = \{p_t\}_{t=1}^{T}, \qquad p_t \in \mathbb{R}^{3}.
\end{equation}
For evaluation, the trajectory is represented as an ordered waypoint sequence and projected
onto the top-down view for judge scoring. The projection is used only for standardized visual
comparison; the intended interpretation remains a path in the underlying 3D environment.

A valid trajectory should satisfy three properties:
\begin{itemize}
    \item it should begin near the intended start region,
    \item it should avoid obvious collisions or implausible shortcuts through obstacles, and
    \item it should terminate at a location compatible with the requested action.
\end{itemize}

\section{Prompt Templates}
\label{app:prompts}

\subsection{Instruction-generation prompt format}
\label{app:prompt_instruction_generation}

\begin{Verbatim}
### Role
You are an expert Embodied-Agent Instruction Generator. Your goal is to analyze visual inputs of a 3D environment and generate precise, executable tasks for a robotic agent.
### Inputs
I have provided two images of the same environment:
1. Image 1: Oblique View (Perspective)
2. Image 2: Top-Down View (Map)
### Critical Pre-Condition
Analyze the Top-Down View first.
If the Top-Down view is incomplete, obstructed, or not clearly visible, you must ignore all other instructions and output ONLY this exact phrase:
"The view is not clear to generate instructions"
---
### Phase 1: Environment Analysis
If the views are clear, analyze both images to construct a mental model of the scene:
- Use the Top-Down View for: Global layout, spatial relationships, distances, and valid paths.
- Use the Oblique View for: Object identification, visual attributes, affordances (what can be opened/lifted), and accessibility.
- Identify: All visible, interactable objects (e.g., furniture, appliances, small items).
### Phase 2: Task Generation
Generate exactly {num_per_level} tasks for EACH of the 3 complexity levels defined below (Total tasks: {num_per_level * 3}).
Task Levels:
- LEVEL_1 (Navigation): Movement from Object A to Object B. No manipulation.
    Constraint: INTERACT must be "none".
- LEVEL_2 (Simple Interaction):* Interaction with 1-2 specific objects.
    Constraint: INTERACT list must contain 1-2 objects.
- LEVEL_3 (Complex Interaction):* Interaction sequences involving 3+ objects or states.
    Constraint: INTERACT list must contain 3+ objects.
---
### Strict Constraints & Guardrails
1.  Object Grounding:
    - Every START and END location must refer to a specific, visible object found in the images.
    - Correct: "Start: Near the red chair"
    - Incorrect: "Start: Near the wall" or "Start: At the start point"
2.  Forbidden Terminology (Spatial Hallucination):
    - NEVER use viewpoint-dependent or relative directional terms.
    - BANNED WORDS: left, right, front, back, center, centre, middle, top, bottom, upper, lower.
    - Tasks must be valid regardless of the agent's facing direction.
3.  Object Consistency:
    - Do not hallucinate objects. Only reference items clearly visible in the provided images.
---
### Output Format
Step 1: Provide an *ENVIRONMENT SUMMARY* (2-3 sentences describing the room type and listing 10-15 key visible objects).
Step 2: Output the task list following this exact schema:
LEVEL_1 | TASK: <instruction> | START: Near <object> | END: Near <object> | INTERACT: none
LEVEL_2 | TASK: <instruction> | START: Near <object> | END: Near <object> | INTERACT: <obj1>, <obj2>
LEVEL_3 | TASK: <instruction> | START: Near <object> | END: Near <object> | INTERACT: <obj1>, <obj2>, <obj3>
### One-Shot Example
Use this structure as a template:
ENVIRONMENT SUMMARY:
This is a modern office space featuring a central workstation and a lounge area. Key objects include: desk, ergonomic chair, monitor, keyboard, filing cabinet, bookshelf, cactus pot, whiteboard, leather sofa, glass coffee table, floor lamp, laser printer, window blinds, and office door.
LEVEL_1 | TASK: Navigate from the entrance to the work desk | START: Near the office door | END: Near the desk | INTERACT: none
LEVEL_1 | TASK: Move from the storage area to the seating zone | START: Near the filing cabinet | END: Near the leather sofa | INTERACT: none
LEVEL_1 | TASK: Walk from the workstation to the meeting area | START: Near the desk | END: Near the whiteboard | INTERACT: none
LEVEL_2 | TASK: Pick up the document from the printer and place it on the desk | START: Near the laser printer | END: Near the desk | INTERACT: document, desk
LEVEL_2 | TASK: Retrieve the book from the bookshelf and place it on the coffee table | START: Near the bookshelf | END: Near the glass coffee table | INTERACT: book, coffee table
LEVEL_2 | TASK: Take the pen from the desk and place it near the whiteboard | START: Near the desk | END: Near the whiteboard | INTERACT: pen, whiteboard
LEVEL_3 | TASK: Open the filing cabinet, retrieve a folder, close the cabinet, and place the folder on the desk | START: Near the filing cabinet | END: Near the desk | INTERACT: cabinet door, folder, desk
LEVEL_3 | TASK: Pick up the laptop from the desk, open it, then move to the sofa | START: Near the desk | END: Near the leather sofa | INTERACT: laptop, laptop lid, leather sofa
LEVEL_3 | TASK: Turn on the floor lamp, pick up the book from the shelf, then place it on the desk | START: Near the floor lamp | END: Near the desk | INTERACT: lamp switch, book, desk
Task:
Analyze the provided images and generate the output now.
\end{Verbatim}

\subsection{Trajectory-prediction prompt format}
\label{app:prompt_trajectory}

\begin{Verbatim}
You are given:
- An oblique view of the same scene
- A top-down view of a 3D scene

Task: Compute the shortest valid walking path while avoiding obstacles on the top-down view.

============================================================
GRID

The top-down view is scene with a discrete 25×25×25 grid.

Valid coordinates:
0 <= x,y,z <= 24
Integers only.
Do NOT use pixel or continuous values.

Origin: (0,0,0) = bottom-left-front floor corner
Positive X axis goes right, positive Y axis goes up, positive Z axis goes inside the screen

Floor = XZ plane.
Estimate (x, z) positions using the top-down view.
Estimate the correct floor height (y) using the oblique view.
All path points must lie on the walkable floor plane.

============================================================
CONSTRAINTS

- All the coordinates should lie on the top-down view
- Determine the ranges of all obstacles (furniture, walls, vehicles, objects, elevated surfaces) on the top-down view
- None of the coordinates (including the coordinates for the starting point) should intersect or touch any obstacle or lie under any object
- Stay inside grid
- Remain on the floor plane
- Carpets, roads, stairs are walkable

============================================================
OUTPUT (STRICT)

Return a valid json of the following format.

Format:
[
  {{"point": [x0, y0, z0], "label": ""}},
  {{"point": [x1, y1, z1], "label": ""}},
  ...
  {{"point": [xN, yN, zN], "label": ""}}
]
\end{Verbatim}

\subsection{Judge prompt format}
\label{app:prompt_judge}

\begin{Verbatim}
You are a visual navigation judge trained to evaluate instruction-guided trajectories in 3D environments.
You must ground all judgments strictly in visible evidence from the images.
Do not assume intent or hidden scene structure.
If evidence is insufficient, explicitly state that it is unclear.

====================
NAVIGATION TASK
====================
{task}

====================
INPUTS
====================
- Image 1: Oblique view of the scene.
- Image 2: Top-down view of the same scene.
- The predicted trajectory is shown using green stars and green connecting lines.
- The green dot marks the start of the trajectory.
- The red dot marks the end of the trajectory.

====================
OUTPUT FORMAT (JSON ONLY)
====================
{{
  start_location_accuracy: {{
    score: 0-5 or N/A,
    justification: string
  }},
  goal_completion: {{
    score: 0-5 or N/A,
    justification: string
  }},
  obstacle_avoidance: {{
    score: 0-5 or N/A,
    justification: string
  }},
  trajectory_efficiency: {{
    score: 0-5 or N/A,
    justification: string
  }},
  overall_summary: string
}}
\end{Verbatim}

\subsection{Judge rubric}
The following table shows the judge rubric that was used to score each predicted trajectory.

\label{app:judge_rubric}
\begin{table}[htbp]
\centering
\small
\caption{Pointwise judge rubric used to score each predicted trajectory.}
\label{tab:judge_rubric}
\begin{tabular}{p{0.24\linewidth}p{0.68\linewidth}}
\toprule
Factor & Definition \\
\midrule
Start Location Consistency & Whether the predicted path begins in the correct initial region or near the intended starting landmark. \\
Goal Satisfaction & Whether the trajectory ends at a location that satisfies the instruction and is compatible with the requested action. \\
Obstacle Avoidance & Whether the route avoids obvious collisions, barrier crossings, or geometrically implausible shortcuts. \\
Trajectory Efficiency & Whether the route is reasonably direct and avoids unnecessary detours relative to the stated objective. \\
\bottomrule
\end{tabular}
\end{table}

\section{Experimental Settings and Reproducibility}
\label{app:reproducibility}

All experiments are conducted in a zero-shot evaluation-only regime. No model is fine-tuned,
adapted, or exposed to benchmark-specific gradient updates. Each model receives the same
task inputs: the same rendered views, the same natural-language instruction, and the same
output formatting requirement.

\subsection{Evaluated models}
\label{app:models}

\begin{table}[htbp]
\centering
\small
\caption{Models evaluated in the current paper.}
\label{tab:evaluated_models}
\begin{tabular}{p{0.28\linewidth}p{0.28\linewidth}p{0.20\linewidth}p{0.14\linewidth}}
\toprule
Model & Inputs & Output & Adaptation \\
\midrule
Qwen3-VL & Top-down view, oblique view, instruction & 3D waypoint sequence & None \\
Gemini Robotics ER-1.5 & Top-down view, oblique view, instruction & 3D waypoint sequence & None \\
GPT-5-mini & Top-down view, oblique view, instruction & 3D waypoint sequence & None \\
\bottomrule
\end{tabular}
\end{table}

\subsection{Decoding and scoring protocol}
\label{app:decoding}

% Each model produces one trajectory per scene--instruction pair under deterministic decoding.
% When the serving API exposes a temperature parameter, we set it to zero; otherwise, we use
% the provider's deterministic or lowest-variance setting. The judge model is fixed to
% \judgemodel{} with temperature set to zero throughout.

Let $\bar{s}_{k}(\tau) \in [0,1]$ denote the normalized score assigned to trajectory $\tau$ on factor
$k \in \{\text{start, goal, obs, eff}\}$. We compute tier-level factor scores by averaging valid
(non-\texttt{N/A}) normalized scores within each difficulty tier, and we compute overall factor
scores by averaging across tiers with equal tier weight.

For reproducibility, we will be releasing the SLEEPWALK dataset containing the 3D environments and the corresponding instructions upon acceptance.

% \subsection{What to release for reproducibility}
% \label{app:release}

% For full reproducibility, the final artifact release should include:
% \begin{itemize}
%     \item the scene-description list used to construct the benchmark,
%     \item the exact prompt templates used for instruction generation, trajectory prediction,
%     and judge scoring,
%     \item rendered top-down and oblique views for each benchmark scene,
%     \item model outputs in the original waypoint format, and
%     \item the code used to render overlays and aggregate judge scores.
% \end{itemize}

% For anonymous review, these materials should be released through an anonymized artifact
% pipeline or deferred until acceptance.

\section{Additional Qualitative Examples}
\label{app:qualitative}

Figure~\ref{fig:additional_examples} is examples generated by GPT5-mini model across the three \benchmarkname{} difficulty tiers (Easy, Medium and Hard). Below are the corresponding Instructions we've provided to the model:

\begin{figure*}[htbp]
\centering
\includegraphics[width=0.31\textwidth]{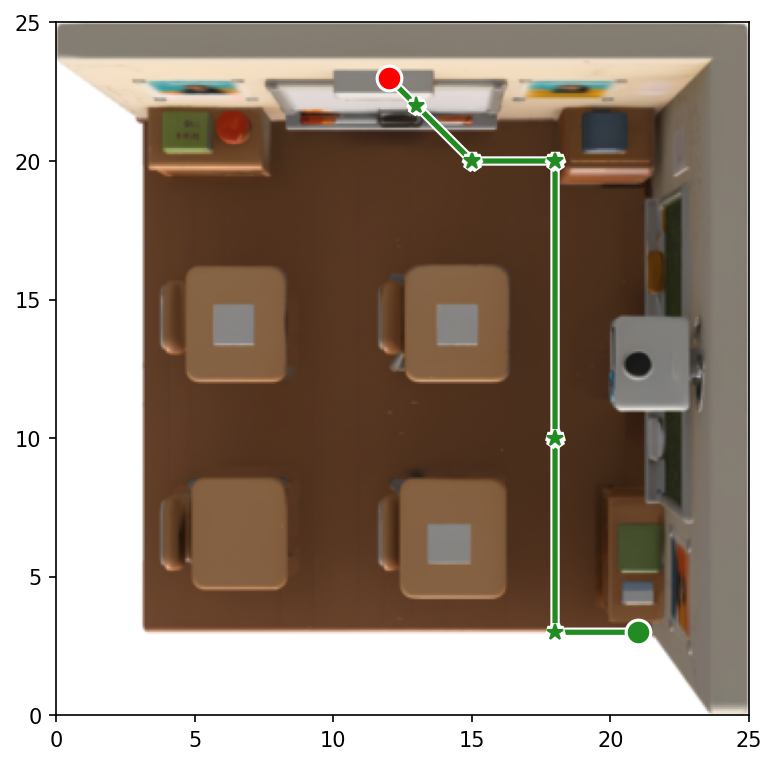}\hfill
\includegraphics[width=0.31\textwidth]{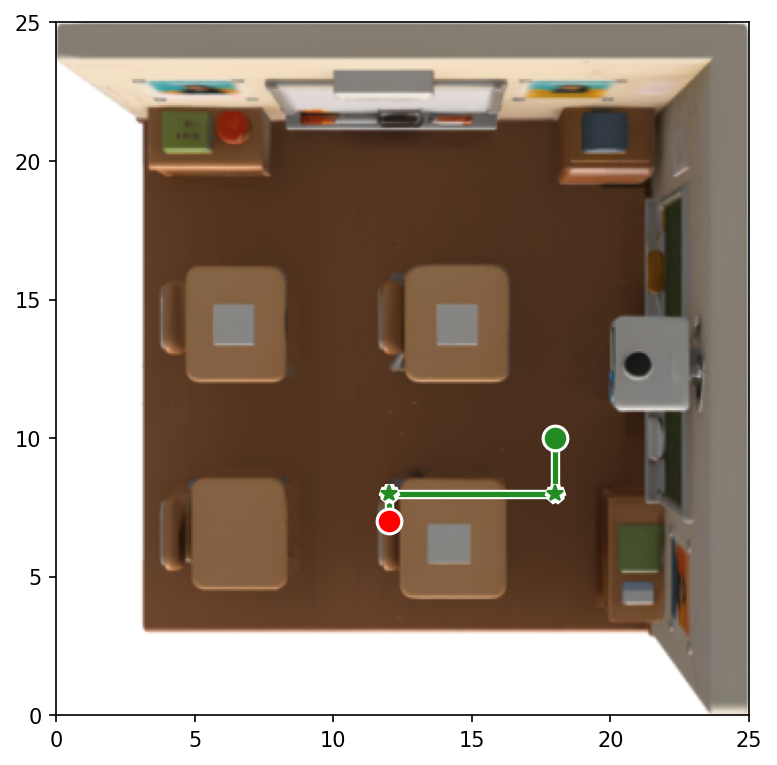}\hfill
\includegraphics[width=0.31\textwidth]{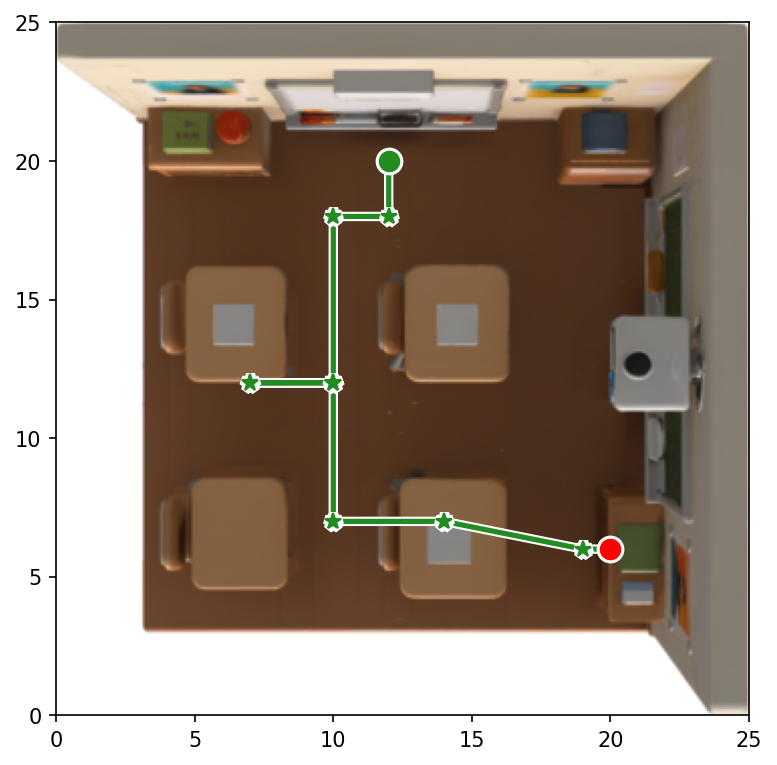}
\caption{Additional qualitative examples generated by GPT5-mini model across the three \benchmarkname{} difficulty tiers (Easy, Medium, Hard).}
\label{fig:additional_examples}
\end{figure*}

\begin{itemize}
    \item LEVEL 1 (Easy) \\TASK: Walk from the trash can to the whiteboard \\START: Near the trash can \\END: Near the whiteboard \\INTERACT: none
    \item LEVEL 2 (Medium) \\TASK: Place the green book from the shelf onto the teacher's desk \\START: Near the shelf \\END: Near the teacher's desk \\INTERACT: green book, teacher's desk
    \item LEVEL 3 (Hard) \\TASK: Move from the whiteboard to the student desk, pick up the book, and place it on the teacher's podium \\START: Near the whiteboard \\END: Near the teacher's podium \\INTERACT: student desk, book, teacher's podium
\end{itemize}

\section{Limitations, Failure Modes, and Ethical Considerations}
\label{app:limitations}

\subsection{Benchmark limitations}
\label{app:benchmark_limitations}

\benchmarkname{} isolates a useful intermediate capability between perception and action,
but it does not fully solve embodied evaluation. First, the benchmark uses reconstructed
single-scene environments rather than fully interactive physics-based worlds, so object
dynamics and low-level contact mechanics are only approximately reflected. Second, the
primary evaluation protocol relies on a strong judge model rather than human annotation
for every sample, which may introduce scoring bias despite the use of explicit rubrics.
Third, top-down trajectory overlays simplify comparison, but they do not capture every
detail of embodied execution.

\subsection{Observed failure modes}
\label{app:failure_modes}

Across models, we repeatedly observe three broad failure patterns:
\begin{itemize}
    \item \textbf{Mislocalized starts:} the model begins from the wrong region even when the
    overall target object is correctly identified.
    \item \textbf{Incorrect or partial goal grounding:} the path is plausible but ends at the wrong
    object, the wrong side of the object, or a location that is not interaction-compatible.
    \item \textbf{Geometrically unsafe execution:} the predicted path appears semantically sensible
    but passes through clutter, clips furniture, or stops in a way that a humanoid agent
    could not realistically execute.
\end{itemize}

\subsection{Ethical considerations}
\label{app:ethics}

This benchmark is intended to support safer and more reliable embodied multimodal systems
by diagnosing failures in grounded spatial reasoning before real-world deployment. At the
same time, better language-conditioned navigation can improve autonomous capabilities in
ways that may be dual-use. %We therefore recommend controlled artifact release, careful
% documentation of benchmark scope, and explicit separation between benchmark performance
% and claims of real-world deployment readiness.

%\input{4_aavi}

\end{document}